\documentclass[]{sensenova}

\usepackage[utf8]{inputenc}
\usepackage[T1]{fontenc}
\usepackage{CJKutf8}

\usepackage[table]{xcolor}
\usepackage{array}
\usepackage{booktabs}
\usepackage{tabularx}
\usepackage{multirow}
\usepackage{makecell}
\usepackage{colortbl}
\usepackage{adjustbox}

\usepackage{graphicx}
\usepackage{wrapfig}
\usepackage{subcaption}
\usepackage{caption}
\usepackage{placeins}

\usepackage{algorithm}
\usepackage{algorithmic}
\usepackage{listings}

\usepackage{soul}
\usepackage{dsfont}
\usepackage{fontawesome5}
\usepackage[misc]{ifsym}
\usepackage{xspace}
\usepackage{enumitem}
\usepackage{verbatim}
\usepackage{framed}
\usepackage{cuted}

\usepackage[toc,page,header]{appendix}

\usepackage{tikz}

\usepackage{mmstyles}

\usepackage{hyperref}
\usepackage{cleveref}

\newcommand{\circnum}[1]{\tikz[baseline=(char.south)]{\node[shape=circle,fill=black,text=white,inner sep=0pt,minimum size=0.8em,font=\scriptsize\rmfamily\bfseries] (char) {\scriptsize #1};}}

\newcommand{\markissue}[2]{\hypertarget{#1}{\circnum{#2}}}
\newcommand{\refissue}[2]{\hyperlink{#1}{\circnum{#2}}}

\newcounter{rq}
\newcommand{\researchquestion}[2]{%
  \refstepcounter{rq}%
  \noindent\textbf{RQ\therq}: #1%
  \label{#2}%
}

\newcommand{\rqref}[1]{\hyperref[#1]{\textbf{RQ\ref*{#1}}}}

\title{AtlasVA: Self-Evolving Visual Skill Memory for Teacher-Free VLM Agents} 

\author{
Pan Wang$^{*,1,2}$, Yihao Hu$^{*,1,3}$, Xiujin Liu$^{4}$, Jingchu Yang$^{1}$, Hang Wang$^{5}$, Zhihao Wen$^{\textrm{\Letter},1}$

\parbox{\textwidth}{\centering\small
    $^{*}$ Equal Contribution \quad
    $\textrm{\Letter}$ Corresponding Author (\href{mailto:z.wen@antgroup.com}{z.wen@antgroup.com}) \\
    $^{1}$ Ant Group \quad
    $^{2}$ University of Science and Technology of China \quad 
    $^{3}$ Westlake University \quad \\
    $^{4}$ University of Michigan - Ann Arbor \quad
    $^{5}$ Sun Yat-sen University \quad
    }}
\abstract{

Vision-language model (VLM) agents increasingly rely on memory-augmented reinforcement learning to reuse experience across long-horizon tasks, yet most existing frameworks store memory as text and depend on proprietary teacher models to summarize or refine it. This design is poorly matched to spatial decision making: geometric priors are compressed into lossy language, and sparse interaction is often supervised through delayed textual feedback rather than dense visually grounded signals. We argue that reusable experience for VLM agents should remain visually grounded. Based on this insight, we propose \textbf{AtlasVA}, a teacher-free visual skill memory framework that organizes memory into three complementary layers: spatial heatmaps, visual exemplars, and symbolic text skills. AtlasVA further evolves danger and affinity atlases directly from trajectory statistics and lightweight grid heuristics, and reuses these self-evolving atlases as potential-based shaping rewards for reinforcement learning. This unifies perception, memory, and optimization without external LLM supervision. Experiments on \textsc{Sokoban}, \textsc{FrozenLake}, 3D embodied navigation, and 3D robotic manipulation benchmarks show that AtlasVA consistently outperforms text-centric memory baselines and competitive VLM agents, with especially strong gains on spatially intensive tasks.
\vspace{-10pt}
\checkdata[Homepage]{\url{https://wangpan-ustc.github.io/AtlasvaWeb/}}
}

\begin{document}
\maketitle

\abstract{

Vision-language model (VLM) agents increasingly rely on memory-augmented reinforcement learning to reuse experience across long-horizon tasks, yet most existing frameworks store memory as text and depend on proprietary teacher models to summarize or refine it. This design is poorly matched to spatial decision making: geometric priors are compressed into lossy language, and sparse interaction is often supervised through delayed textual feedback rather than dense visually grounded signals. We argue that reusable experience for VLM agents should remain visually grounded. Based on this insight, we propose \textbf{AtlasVA}, a teacher-free visual skill memory framework that organizes memory into three complementary layers: spatial heatmaps, visual exemplars, and symbolic text skills. AtlasVA further evolves danger and affinity atlases directly from trajectory statistics and lightweight grid heuristics, and reuses these self-evolving atlases as potential-based shaping rewards for reinforcement learning. This unifies perception, memory, and optimization without external LLM supervision. Experiments on \textsc{Sokoban}, \textsc{FrozenLake}, 3D embodied navigation, and 3D robotic manipulation benchmarks show that AtlasVA consistently outperforms text-centric memory baselines and competitive VLM agents, with especially strong gains on spatially intensive tasks.
\vspace{-10pt}
\checkdata[Homepage]{\url{https://wangpan-ustc.github.io/AtlasvaWeb/}}
}

\section{Introduction}
\label{sec:intro}

Vision-language models (VLMs) are becoming a practical interface for interactive agents that must read instructions, parse screenshots, and execute grounded actions in sequential environments~\cite{niu2024screenagent,lin2025showui,gou2025navigating,wu2025gui,zhou2025gui,luo2025visual}. This setting is especially challenging in long-horizon spatial tasks, where the agent must accumulate reusable experience rather than reason from scratch at every step. Recent memory-augmented reinforcement learning (RL) systems address this need by storing retrieved skills, retrospectives, and task heuristics across episodes~\cite{shinn2023reflexion,zhang2023large,sridhar2025regent,xu2025mem}. However, most of these systems still treat the VLM agent as a text-centric reasoner: the visual stream is consumed only as a transient observation, while reusable knowledge is stored almost entirely in language~\cite{deng2023mind2web,koh2024visualwebarena}.

Recent memory-centric frameworks show that explicit skill libraries can improve exploration and reuse past experience, with representative examples such as SkillRL~\cite{xia2026skillrl} and XSkill~\cite{jiang2026xskill}. Their common recipe is to summarize successful or failed trajectories into text, retrieve the relevant summaries for a new state, and then rely on a strong teacher model to refine the memory over time. This recipe is reasonable for language-only agents, but it is poorly matched to VLM agents that solve tasks through spatial perception~\cite{10495141,10938647,liu2024volumetric}. Compressing a two-dimensional layout into one-dimensional textual rules discards geometric structure, turning visually grounded decision making into a lossy translation problem.

\begin{wrapfigure}{l}{0.48\linewidth}
  \centering
  \includegraphics[width=0.98\linewidth]{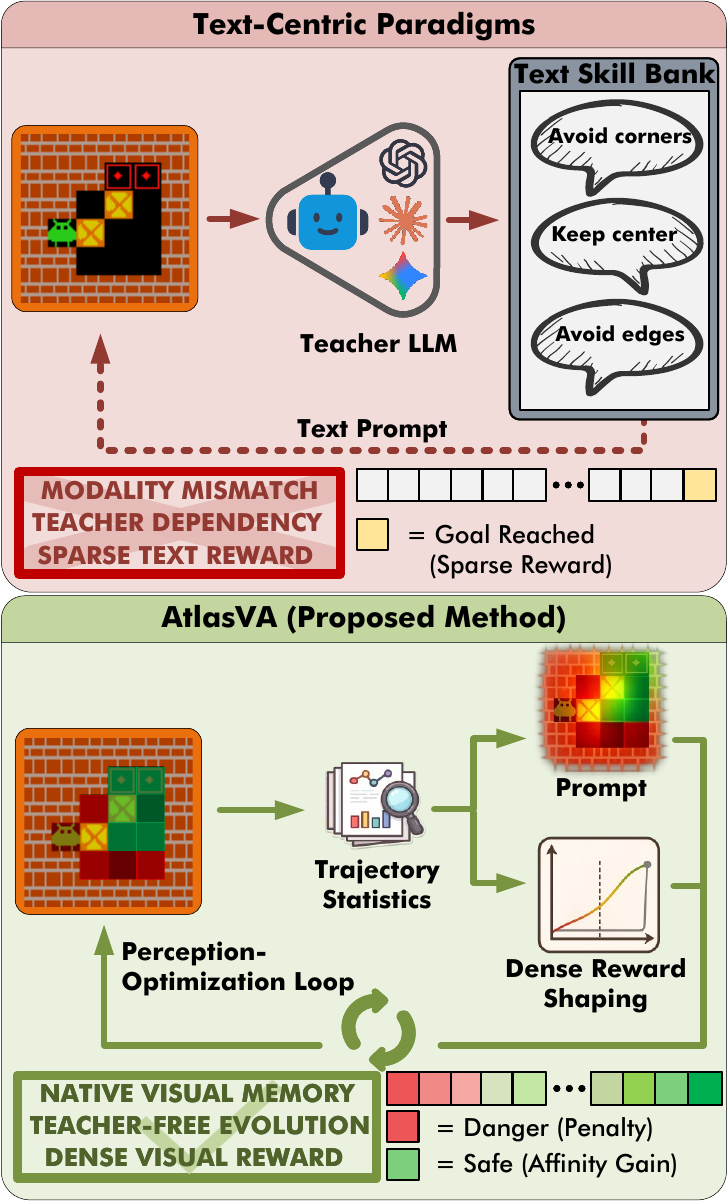}
  \caption{Comparison between text-centric memory paradigms and \textbf{AtlasVA}. }
  \label{fig:teaser_fig}
  \vspace{-5pt}
\end{wrapfigure}

Under this text-centric design, VLM agents face unique challenges when interacting with complex environments, as illustrated in Figure~\ref{fig:teaser_fig} (top).  
Primarily, \markissue{issue:modality}{1} \textit{encoding high-value spatial cues into text inevitably incurs severe information loss}, as text cannot capture rich visual details. % \begin{figure}[!t]
%   \centering
%   \includegraphics[width=1.0\linewidth]{figures/Teaser1.pdf}
%   \caption{Comparison between text-centric memory paradigms and \textbf{AtlasVA}.}
%   \label{fig:teaser_fig}
% \end{figure}

% \begin{figure}[!t]
%   \centering
%   \includegraphics[width=1.0\linewidth]{figures/Teaser1.pdf}
%   \caption{Comparison between text-centric memory paradigms and \textbf{AtlasVA}.While existing frameworks compress spatial experience into teacher-generated text, causing modality mismatch and sparse textual feedback, AtlasVA maintains native visual memory, evolves spatial priors from trajectory statistics, and uses them for dense visual reward shaping in a unified perception-optimization loop.}
%   \label{fig:teaser_fig}
% \end{figure}

 For example, text struggles to convey complex spatial topologies, such as dead ends, local hazards, and promising sub-goal regions. 
Furthermore, to manage these textual records, \markissue{issue:teacher}{2} \textit{many frameworks require a proprietary LLM to summarize failures, merge skills, or rewrite memory.} This dependence not only increases computational costs but also undermines the premise of fully autonomous self-improvement.
Finally, using textual rewards to evaluate spatial interactions creates a \markissue{issue:feedback}{3} \textit{feedback mismatch for RL}. Because textual critiques are inherently abstract and lack precise spatial grounding, they deprive the agent of dense, coordinate-aware guidance. This exacerbates the credit assignment problem, underscoring the critical need to align the reward format with the spatial nature of the task.

In this paper, we propose \textbf{AtlasVA}, which transitions agent memory from a text-only repository to a \textit{Visual Skill Memory (VSM)} hierarchy (Figure~\ref{fig:teaser_fig}, bottom). \textbf{First}, VSM stores reusable experience across three complementary layers: spatial heatmaps, visual exemplars, and symbolic text to address \refissue{issue:modality}{1}. During forward inference, these heatmaps act as visual prompts that project complex 3D environments into 2.5D spatial maps, effectively maintaining sensitive spatial knowledge within the VLM's native modality. \textbf{Second}, due to \refissue{issue:teacher}{2}, AtlasVA bootstraps memory directly from raw interaction history. Instead of using external models for textual summarization, it aggregates trajectory statistics to construct self-evolving spatial heatmaps, thereby reducing computational overhead and ensuring fully autonomous self-improvement. \textbf{Third}, to resolve \refissue{issue:feedback}{3}, these same heatmaps are mathematically formulated as a potential function for policy updates. By rewarding movement toward historically successful regions and penalizing identified deadlocks, this mechanism provides the dense, coordinate-aware guidance necessary to alleviate credit assignment difficulties.

Overall, our main contributions are as follows:
\vspace{-5pt}
\begin{itemize}
    \setlength{\itemsep}{1pt}
    \setlength{\parskip}{1pt}
    \setlength{\parsep}{1pt}
    \item We propose \textbf{AtlasVA}, a memory framework for VLM agents that replaces lossy text-only storage with a \textbf{Visual Skill Memory} hierarchy. By preserving spatial heatmaps, visual exemplars, and symbolic text, it natively aligns reusable experience with the agent's visual perception.
    \item We introduce a \textbf{teacher-free atlas evolution} mechanism to sustain this memory. This mechanism refines danger and affinity heatmaps directly from raw trajectory statistics, removing the need for proprietary LLM supervision.
    \item We develop a \textbf{dense visual reward shaping} strategy that repurposes the evolving visual priors as potential functions. This provides dense, coordinate-aligned gradients that mitigate the severe sparse-reward bottleneck in both 2D and 3D tasks, yielding substantial gains in both learning efficiency and final task performance.

\end{itemize}

\section{Related Work}
\label{sec:related_works}

\subsection{Vision-Language Models as Autonomous Agents}
Large vision-language models (VLMs) have emerged as capable autonomous agents for sequential decision-making in interactive environments ~\cite{paglieri2024balrog,kang2025gflowvlm}. By combining visual grounding with instruction following, VLM agents achieve strong performance in domains such as web navigation, GUI control, and embodied planning~\cite{koh2024visualwebarena,li2025towards,bai2024digirl,wang2024omnijarvis}. However, most current architectures treat visual observations merely as transient, step-level inputs~\cite{sarch2025grounded}. To maintain historical context or long-horizon heuristics, these systems typically compress past experiences into textual representations, such as verbal summaries~\cite{shinn2023reflexion}, scratchpads~\cite{yao2022react}, or retrieved passages~\cite{asai2023self,sarthi2024raptor}. This reliance on text creates an architectural asymmetry when operating in highly spatial environments, as verbal rules struggle to preserve fine-grained geometric layouts and topological constraints. In contrast, AtlasVA eliminates this text bottleneck by storing and updating spatial structures directly within the visual modality, aligning the agent's memory with its native perception.

\subsection{Memory and Skill Evolution in RL Agents}
To enable iterative self-improvement, recent reinforcement learning (RL) frameworks increasingly equip agents with external memory and skill evolution mechanisms. For instance, Reflexion \cite{shinn2023reflexion} uses verbal self-critique to revise plans, while frameworks like ExpeL \cite{zhao2024expel} and Mem0 \cite{chhikara2025mem0} distill trajectories into retrievable linguistic records. Similarly, skill-augmented pipelines such as SkillRL \cite{xia2026skillrl} and XSkill \cite{jiang2026xskill} accelerate learning by constructing libraries of natural-language skills or Markdown workflows from interaction logs. These approaches, however, share two critical limitations: a strict reliance on text-based representations and strong teacher dependence. Even visually grounded frameworks like XSkill store experiences as Markdown or JSON, which inherently struggle to capture fine-grained spatial hazards, topological dead ends, and geometric traps~\cite{cui2024frontier,cheng2024spatialrgpt,yang2025thinking,wang2025mp}. Furthermore, generating and refining these textual rules requires repeated calls to powerful, proprietary large language models, driving up API costs and limiting the agent's true autonomy~\cite{wang2025efficient}. AtlasVA circumvents these bottlenecks by shifting online adaptation entirely to \textit{visual} atlases, which evolve directly from trajectory statistics without requiring external teacher supervision.

\subsection{Reward Shaping and Visual vs. Textual Feedback}
Sparse terminal rewards remain a fundamental bottleneck in RL, often leading to extreme sample inefficiency in spatial reasoning tasks~\cite{feng2025rewardmap,jiang2025episodic}. To address this in VLM agents, recent methods prompt external LLMs with action logs to generate ``textual rewards'' (e.g., ``you should not have pushed the box to the corner'')~\cite{zhong2024policy,wei2025gtr,yang2024text2reward,rocamonde2023vision,wang2024rl}. More formally, potential-based reward shaping (PBRS) \cite{ng1999policy} offers a rigorous solution by adding the difference of a potential function $\Phi$ to the reward. However, both paradigms face significant challenges in visual domains. Textual critiques are abstract and delayed, failing to translate into the dense, coordinate-aligned feedback necessary for spatial optimization, leaving agents to rely on blind trial-and-error when stuck in topological traps. Meanwhile, defining effective PBRS potentials typically requires heavy manual engineering (e.g., hardcoded distance heuristics or static penalty zones) that fails to scale to diverse or dynamic layouts. Standard exploration bonuses like count-based visitation \cite{bellemare2016unifying} also struggle to scale directly to pixel-level visual inputs. AtlasVA bridges this gap by introducing a \textbf{self-evolving visual atlas} that acts as an automatically learned potential function. By combining grid cues with trajectory statistics, it converts spatial danger and affinity maps directly into dense, coordinate-aligned shaping rewards, automating reward engineering while maintaining precise spatial gradients at every step.
\section{AtlasVA}
\label{sec:AtlasVA}

%\caption{\textbf{Overview of the AtlasVA architecture.} The framework operates in a closed perception-optimization loop. \textbf{Left:} The Visual Skill Memory (VSM) prompts the VLM policy using three layers: spatial heatmaps, visual exemplars, and text skills. \textbf{Top Right:} During rollouts, trajectory logs and screenshots drive a teacher-free evolution to update the VSM via EMA blending and exemplar mining. \textbf{Bottom:} The evolved spatial priors are converted into affinity gains and danger penalties for dense reward shaping, guiding policy optimization and closing the loop.}

\begin{figure}[t]
  \centering
  \includegraphics[width=1.0\linewidth]{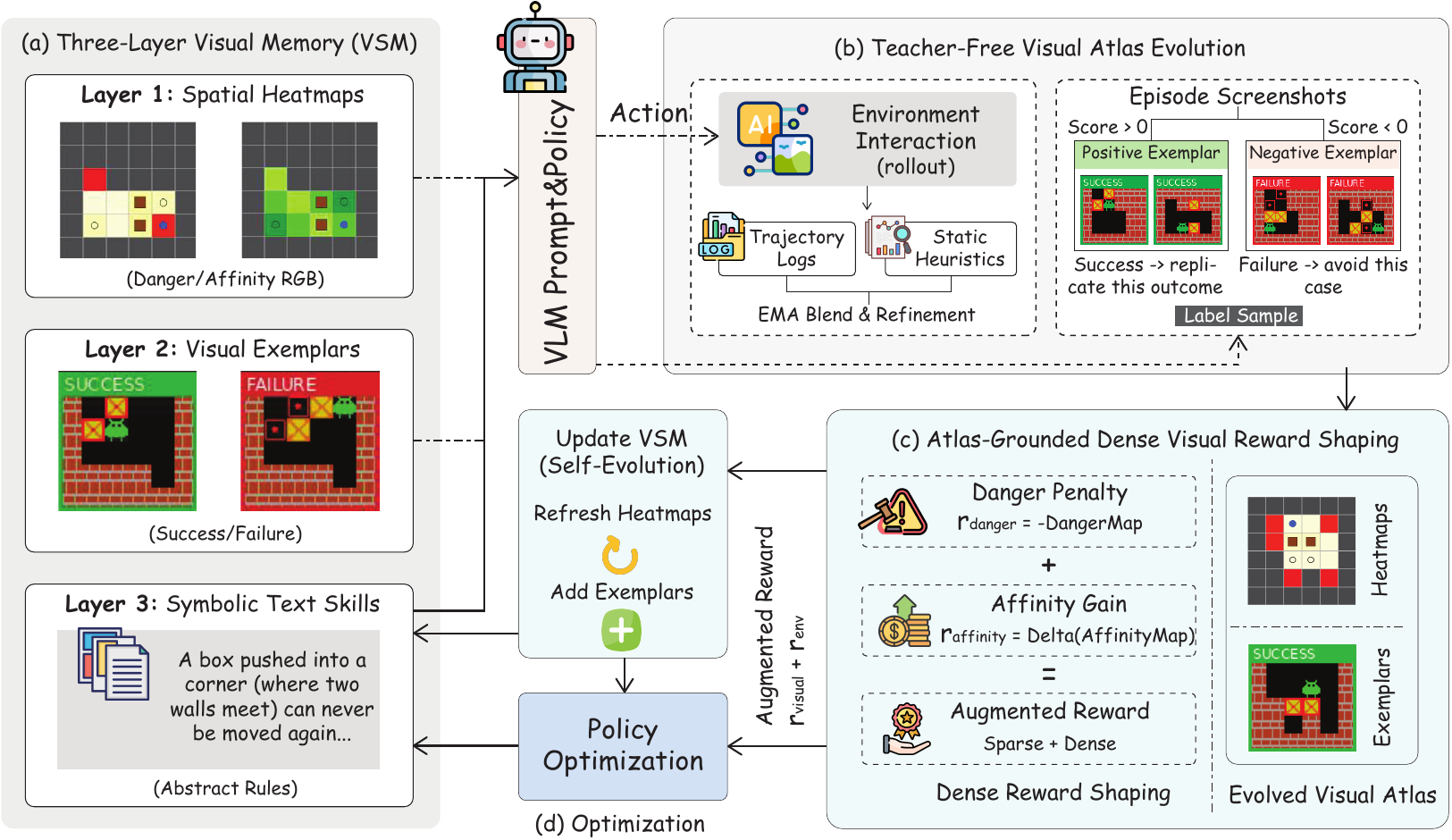}
    \caption{\textbf{Overview of the AtlasVA architecture.} (a) The Visual Skill Memory (VSM) prompts the policy using a three-layer hierarchy: spatial heatmaps, visual exemplars, and symbolic text rules. (b) Through teacher-free environment rollouts, trajectory logs and episode screenshots drive the self-evolution of the VSM via EMA blending and exemplar mining. (c) These evolved spatial priors are formulated into affinity gains and danger penalties, providing dense reward shaping. (d) In the optimization phase, the augmented rewards guide policy updates, while improved trajectories continuously refresh the VSM, forming a closed perception-optimization loop.}
  \label{fig:arch}
\end{figure}

%In this section, we present \textbf{AtlasVA}, a self-evolving visual skill memory designed for teacher-free VLM agents to address sparse rewards and spatial blindness in multimodal reinforcement learning. As illustrated in Figure~\ref{fig:arch}, AtlasVA establishes a closed perception-optimization loop. We first formulate the problem setting (Section~\ref{subsec:problem}). We then introduce the Three-Layer Visual Skill Memory that grounds agent perception in multimodal spatial context (Section~\ref{subsec:vsm}). Next, we detail the Teacher-Free Visual Atlas Evolution mechanism that bootstraps spatial priors from trajectory statistics without external LLMs (Section~\ref{subsec:evolution}). We then explain how the evolved atlas functions as a dense potential function for reward shaping (Section~\ref{subsec:reward}). Finally, we describe the unified prompt interface for injecting these memory components (Section~\ref{subsec:injection}).

\subsection{Problem Formulation}
\label{subsec:problem}

We formulate the embodied decision-making process as a partially observable Markov decision process $\mathcal{M} = \langle \mathcal{S}, \mathcal{O}, \mathcal{A}, \mathcal{T}, \mathcal{R}, \gamma \rangle$. At each timestep $t$, the agent receives a multimodal observation $o_t \in \mathcal{O}$ comprising a rendered RGB frame and a natural language task description. The agent emits a discrete action $a_t \in \mathcal{A}$ sampled from a multimodal policy $\pi_\theta(a_t \mid o_{\le t})$, parameterized by a vision-language model. The environment transitions to a new hidden state $s_{t+1}$ according to the transition dynamics $\mathcal{T}(s_{t+1} \mid s_t, a_t)$ and yields a sparse binary reward $\mathcal{R}(s_t, a_t) \in \{0, 1\}$ upon task success. 
The primary challenge in this setting is that the optimal policy requires profound spatial reasoning and long-horizon planning, yet the environment only provides sparse, delayed feedback. Standard VLM agents struggle to extract persistent topological priors from transient observations. To address this, we define a unified \textit{GridState} abstraction $g_t$ extracted directly from the simulator's internal state to represent semantic layers, including obstacles and interactive objects~\cite{jiang2024visual}. Let $\mathbf{p}_t \in \mathbb{Z}^2$ denote the 2D coordinate of the primary manipulated entity at timestep $t$. AtlasVA leverages this privileged representation exclusively during training to construct and evolve persistent spatial priors across episodes; the deployed VLM policy relies solely on visual inputs (see Appendix~\ref{sec:supp_gridstate_api} for API details).

\subsection{Three-Layer Visual Skill Memory (VSM)}
\label{subsec:vsm}
(\textbf{Fig.~\ref{fig:arch}a}) To rectify the modality mismatch of text-centric rules, we introduce a multimodal memory architecture comprising three complementary layers. The augmented multimodal prompt provided to the VLM is constructed as $\tilde{o}_t = \left[ M_{danger}, M_{affinity}, \mathcal{E}_{vis}, \mathcal{S}_{text}, o_t \right]$. 

\textbf{Layer 1: Spatial Heatmaps.} At the foundational level, we provide continuous spatial fields over the environment: a \textit{danger map} indicating deadlock risks and an \textit{affinity map} indicating proximity to task completion. Rather than encoding these as text arrays, we render them as RGB heatmaps $M \in \mathbb{R}^{H \times W \times 3}$, which natively align with the VLM visual encoder. These are injected as separate visual tokens or alpha-blended over the original observation $o_t$, allowing the agent to intuitively perceive hazardous and promising regions without modality translation. 

\textbf{Layer 2: Visual Exemplars.} To complement these abstract spatial priors with concrete visual references, the second layer introduces visual exemplars ($\mathcal{E}_{vis}$) by mining a small bank of representative screenshots from historical rollouts, explicitly annotated with success or failure tags. 

\textbf{Layer 3: Symbolic Text Skills.} Finally, to preserve high-level strategic reasoning, the third layer integrates symbolic text skills ($\mathcal{S}_{text}$), retaining a compact set of textual heuristics. Together, these components establish a complete knowledge gradient from perceptual spatial maps and concrete layout examples to symbolic logic, providing holistic, in-context guidance without the lossy translation inherent in purely text-based memory.

\subsection{Teacher-Free Visual Atlas Evolution}
\label{subsec:evolution}

(\textbf{Fig.~\ref{fig:arch}b}) To eliminate reliance on external LLM teachers, we propose a purely data-driven forward process that bootstraps spatial priors directly from the agent's own interaction data. The spatial heatmaps in Layer 1 are synthesized by fusing static grid heuristics $M_{heuristic}$ with accumulated trajectory statistics $M_{stat}$. 

\textbf{Trajectory accumulation.} First, for the static heuristics branch, we extract topological features from the current layout, such as corner-like deadlock regions and Breadth-First Search distances to target objects. Second, for the trajectory statistics branch, we analyze the structured rollout logs of the current training batch. Let $\mathcal{T}_{fail}$ and $\mathcal{T}_{succ}$ denote the set of failed and successful trajectories. We compute the batch-level danger map by accumulating the terminal failure positions $\mathbf{p}_T$:

\begin{equation}
    M_{batch}^{danger}(\mathbf{p}) = \frac{1}{|\mathcal{T}_{fail}|} \sum_{\tau \in \mathcal{T}_{fail}} \mathbb{I}(\mathbf{p}_T = \mathbf{p})
\end{equation}
Similarly, the batch-level affinity map $M_{batch}^{affinity}$ is derived by recording the normalized visit frequency of coordinates along the successful paths in $\mathcal{T}_{succ}$.

\textbf{EMA blending.} Next, we update the historical trajectory statistics using an Exponential Moving Average (EMA) with decay rate $\alpha$: $M_{stat} \leftarrow \alpha M_{stat} + (1 - \alpha) M_{batch}$. Finally, the output heatmap presented to the VLM is a dynamic blend of the static heuristics and the EMA statistics:
\begin{equation}
    M_{final} = (1 - \beta_k) M_{heuristic} + \beta_k M_{stat}
\end{equation}
where $\beta_k \in [0, 1]$ is a scheduling coefficient that progressively anneals from $0$ to $1$ over training epochs $k$. This scheduling coefficient provides safe, cold-start exploration guided by static geometry in early stages, while smoothly transitioning to experience-driven refinement in later stages without risking catastrophic forgetting.

\subsection{Atlas-Grounded Dense Visual Reward Shaping}
\label{subsec:reward}

(\textbf{Fig.~\ref{fig:arch}c}) To alleviate extreme sample inefficiency in sparse reward settings, we design an automated reward shaping mechanism that utilizes our self-evolving spatial atlas as a dynamic potential function. Given a transition from $\mathbf{p}_t$ to $\mathbf{p}_{t+1}$, we compute a bounded auxiliary visual reward $r_{visual} = \lambda_{danger} \cdot r_{danger} + \lambda_{affinity} \cdot r_{affinity}$.

\textbf{Danger penalty.} First, the danger penalty $r_{danger}$ discourages the agent from exploring regions that the atlas has historically identified as risky. We apply a negative penalty proportional to the danger value of the coordinate the primary entity enters: $r_{danger} = -M_{final}^{danger}(\mathbf{p}_{t+1})$. 

\textbf{Affinity gain.} Second, the affinity gain $r_{affinity}$ incentivizes the agent to follow paths that move closer to the goal manifold. We formulate it as the per-step difference of the affinity potential:
\begin{equation}
    r_{affinity} = M_{final}^{affinity}(\mathbf{p}_{t+1}) - M_{final}^{affinity}(\mathbf{p}_{t})
\end{equation}
Because $M_{final}^{affinity}$ inherits the BFS distance gradient from $M_{heuristic}^{affinity}$, this difference produces a strictly positive signal whenever the agent moves one step closer to the goal, and a strictly negative signal when it moves away.

\subsection{Optimization and Closed Loop}
\label{subsec:optimization}

\begin{figure}[ht]
    \centering
    \includegraphics[width=0.9\linewidth]{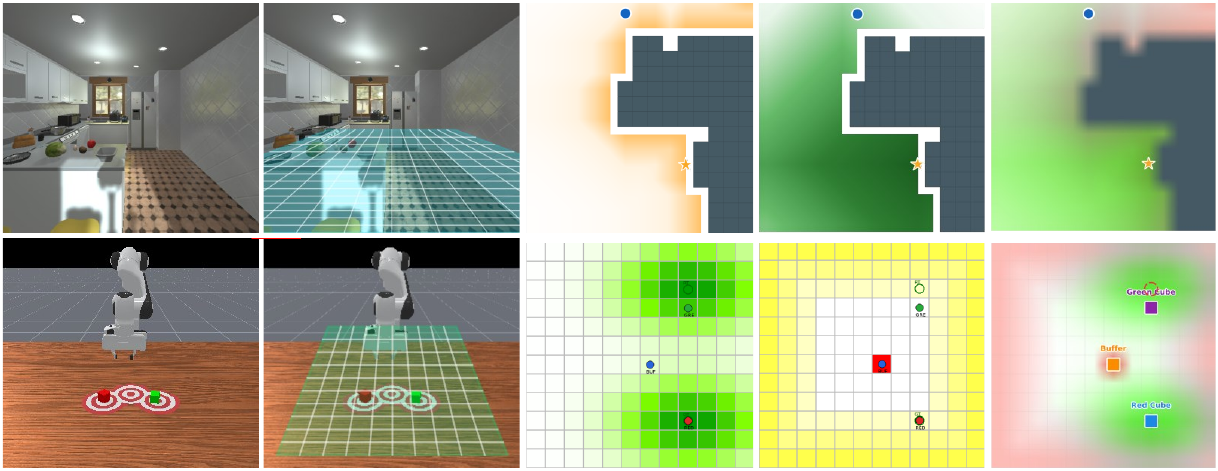}
    \caption{\textbf{Projection of 3D continuous spaces into 2.5D visual priors.} \textbf{Top row:} In Navigation, 3D rooms are mapped to 2D floor plans with obstacle-aware heatmaps. \textbf{Bottom row:} In PrimitiveSkill, tabletop workspaces are discretized into 2.5D grids, separating x-y planar guidance from continuous z-axis constraints.}
    \label{fig:3d_projection}
\end{figure}

(\textbf{Fig.~\ref{fig:arch}d}) The final reward used for RL optimization is $r_{visual} + r_{env}$. While the affinity gain is formulated as a potential difference~\cite{ng1999policy}, the danger penalty acts as a heuristic safety constraint that intentionally alters the optimal policy to prioritize safe navigation over hazardous shortcuts. Furthermore, this mechanism closes the perception-optimization loop autonomously: improved policies generate higher-quality trajectories, which refine the visual atlas via EMA, which in turn provides more accurate dense rewards for further optimization (see Figure~\ref{fig:3d_projection} and Appendix~\ref{sec:supp_3d_projection} for details on projecting 3D continuous spaces into 2.5D visual priors).

To prevent modality mismatch when injecting RGB heatmaps, visual exemplars, and symbolic text back into the policy during this loop, AtlasVA establishes a unified, interleaved vision-language prompt interface. As summarized in Table~\ref{tab:vsm_injection}, each VSM layer is assigned a distinct modality, update rule, and prompt anchor. We natively inject the spatial heatmaps (\textbf{Layer 1}) and episodic visual exemplars (\textbf{Layer 2}) as standalone visual tokens (\textlangle image\textrangle). These visual priors are hierarchically anchored alongside the symbolic text rules (\textbf{Layer 3}) as shown in the prompt skeleton (Appendix~\ref{sec:supp_prompt_template}). This native multimodal injection empowers the VLM to perform zero-shot spatial pattern matching directly in the pixel space. It cleanly bridges the gap between abstract strategic planning and visually-grounded execution, ensuring the VLM leverages its pre-trained visual encoders efficiently.

\begin{table}[h]
\centering
\caption{VSM injection specification. Each layer is assigned a distinct modality, update rule, and anchor position in the VLM prompt, forming a unified visual-textual memory interface.}
\label{tab:vsm_injection}
\small
\resizebox{\linewidth}{!}{
\begin{tabular}{llllc}
\toprule
\textbf{Layer} & \textbf{Modality} & \textbf{Content} & \textbf{Update Rule} & \textbf{Prompt Anchor} \\
\midrule
L1 Heatmap  & RGB image & Danger / Affinity map  & EMA over trajectories      & \texttt{\#\# Spatial Skill Maps} \\
L2 Exemplar & RGB image(s) & Top-$k$ success / failure frames & Retrieval + eviction (cap=6) & Before current observation \\
L3 Text     & Text & Principles \& mistakes & Per-episode summarization  & \texttt{\#\# Learned Principles} \\
\bottomrule
\end{tabular}
}
\end{table}
\section{Experiments}
\label{sec:experiments}
\definecolor{monte_carlo}{RGB}{131, 195, 176}

\newcommand{\numwild}[2]{%
    \begin{tikzpicture}[baseline]
        \pgfmathsetmacro{\range}{1.0 - #2}%
        \pgfmathparse{\range > 0 ? 1 : 0}%
        \ifnum\pgfmathresult=1
            \pgfmathsetmacro{\intensity}{min(130, 130*(#1 - #2)/\range)}%
        \else
            \pgfmathsetmacro{\intensity}{0}%
        \fi
        \pgfmathparse{\intensity > 0.5 ? 1 : 0}%
        \ifnum\pgfmathresult=1
            \fill[monte_carlo!\intensity!white, rounded corners=1] (-0.6em, -0.3em) rectangle (2.6em, 1em);
        \fi
        \node[inner sep=0pt] at (1em, 0.7ex) {#1};
    \end{tikzpicture}%
}

\newcommand{\numref}[2]{%
    \begin{tikzpicture}[baseline]
        \pgfmathsetmacro{\range}{1.0 - #2}%
        \pgfmathparse{\range > 0 ? 1 : 0}%
        \ifnum\pgfmathresult=1
            \pgfmathsetmacro{\intensity}{min(130, 130*(#1 - #2)/\range)}%
        \else
            \pgfmathsetmacro{\intensity}{0}%
        \fi
        \pgfmathparse{\intensity > 0.5 ? 1 : 0}%
        \ifnum\pgfmathresult=1
            \fill[monte_carlo!\intensity!white, rounded corners=1] (-0.6em, -0.3em) rectangle (2.6em, 1em);
        \fi
        \node[inner sep=0pt] at (1em, 0.7ex) {\textbf{#1}};
    \end{tikzpicture}%
}

\newcommand{\colorbarvertical}{%
    \begin{tikzpicture}
        \shade[top color=white, bottom color=monte_carlo!130, rounded corners=0.6] (0, 0) rectangle (0.3, 5.5);
        \foreach \y/\lab in {0/100, 0.55/90, 1.1/80, 1.65/70, 2.2/60, 2.75/50, 3.3/40, 3.85/30, 4.4/20, 4.95/10, 5.5/0} {
            \draw[white, semithick] (0, \y) -- (0.05, \y);
            \draw[white, semithick] (0.25, \y) -- (0.3, \y);
            \node[right, font=\footnotesize] at (0.32, \y) {\lab};
        }
        \node[right, font=\small\bfseries] at (0.1, -0.4) {\%};
    \end{tikzpicture}%
}

\begin{table*}[t]
    \centering
    \small
    \begin{minipage}[c]{0.93\textwidth}
    \resizebox{\linewidth}{!}{
    \begin{tabular}{l|c|c|ccc|ccccc|c}
    \toprule
    \multirow{2}{*}{Model/Method} 
    & \multirow{2}{*}{Sokoban} 
    & \multirow{2}{*}{FrozenLake} 
    & \multicolumn{3}{c|}{Navigation} 
    & \multicolumn{5}{c|}{PrimitiveSkill}
    & \multirow{2}{*}{Overall} \\
    \cmidrule(lr){4-6} \cmidrule(lr){7-11}
    & & 
    & Base & Common & Average 
    & Place & Stack & Drawer & Align & Swap 
    & \\
    \midrule
    \rowcolor[HTML]{DFF1EC} 
    \multicolumn{12}{c}{Proprietary Models} \\
    \midrule
    
    GPT-5~\cite{openai2025gpt5}
    & \numwild{0.70}{0.13} & \numwild{0.77}{0.13}
    & \numwild{0.75}{0.22} & \numwild{0.81}{0.27} & \numwild{0.78}{0.24}
    & \numwild{1.00}{0.00} & \numwild{0.63}{0.00} & \numwild{0.00}{0.00} & \numwild{1.00}{0.00} & \numwild{0.55}{0.00}
    & \numwild{0.69}{0.09} \\
    
    o3~\cite{openai2025o3o4mini}
    & \numwild{0.60}{0.13} & \numwild{0.78}{0.13}
    & \numwild{0.81}{0.22} & \numwild{0.75}{0.27} & \numwild{0.78}{0.24}
    & \numwild{1.00}{0.00} & \numwild{0.63}{0.00} & \numwild{0.00}{0.00} & \numwild{1.00}{0.00} & \numwild{0.82}{0.00}
    & \numwild{0.71}{0.09} \\
    
    o4-mini~\cite{openai2025o3o4mini}
    & \numwild{0.44}{0.13} & \numwild{0.82}{0.13}
    & \numwild{0.75}{0.22} & \numwild{0.75}{0.27} & \numwild{0.75}{0.24}
    & \numwild{1.00}{0.00} & \numwild{0.50}{0.00} & \numwild{0.00}{0.00} & \numwild{0.75}{0.00} & \numwild{0.33}{0.00}
    & \numwild{0.60}{0.09} \\
    
    GPT-4o~\cite{hurst2024gpt}
    & \numwild{0.43}{0.13} & \numwild{0.54}{0.13}
    & \numwild{0.75}{0.22} & \numwild{0.69}{0.27} & \numwild{0.72}{0.24}
    & \numwild{0.50}{0.00} & \numwild{0.63}{0.00} & \numwild{0.00}{0.00} & \numwild{0.88}{0.00} & \numwild{0.94}{0.00}
    & \numwild{0.60}{0.09} \\
    
    Gemini 2.5 flash~\cite{google2025gemini25}
    & \numwild{0.52}{0.13} & \numwild{0.57}{0.13}
    & \numwild{0.58}{0.22} & \numwild{0.56}{0.27} & \numwild{0.57}{0.24}
    & \numwild{0.75}{0.00} & \numwild{0.50}{0.00} & \numwild{0.00}{0.00} & \numwild{0.88}{0.00} & \numwild{0.82}{0.00}
    & \numwild{0.58}{0.09} \\
    
    Gemini 2.5 Pro~\cite{google2025gemini25}
    & \numwild{0.58}{0.13} & \numwild{0.78}{0.13}
    & \numwild{0.63}{0.22} & \numwild{0.63}{0.27} & \numwild{0.63}{0.24}
    & \numwild{0.63}{0.00} & \numwild{0.63}{0.00} & \numwild{0.00}{0.00} & \numwild{0.75}{0.00} & \numwild{0.00}{0.00}
    & \numwild{0.51}{0.09} \\
    
    Gemini 2.0~\cite{team2023gemini}
    & \numwild{0.28}{0.13} & \numwild{0.61}{0.13}
    & \numwild{0.50}{0.22} & \numwild{0.63}{0.27} & \numwild{0.56}{0.24}
    & \numwild{0.75}{0.00} & \numwild{0.13}{0.00} & \numwild{0.00}{0.00} & \numwild{0.25}{0.00} & \numwild{0.33}{0.00}
    & \numwild{0.39}{0.09} \\
    
    Claude Sonnet 4.5~\cite{anthropic2025claudesonnet45}
    & \numwild{0.31}{0.13} & \numwild{0.80}{0.13}
    & \numwild{0.67}{0.22} & \numwild{0.67}{0.27} & \numwild{0.67}{0.24}
    & \numwild{0.63}{0.00} & \numwild{0.50}{0.00} & \numwild{0.00}{0.00} & \numwild{1.00}{0.00} & \numwild{1.00}{0.00}
    & \numwild{0.62}{0.09} \\
    
    Claude Sonnet 3.7~\cite{TheC3}
    & \numwild{0.25}{0.13} & \numwild{0.69}{0.13}
    & \numwild{0.48}{0.22} & \numwild{0.47}{0.27} & \numwild{0.47}{0.24}
    & \numwild{0.63}{0.00} & \numwild{0.13}{0.00} & \numwild{0.00}{0.00} & \numwild{1.00}{0.00} & \numwild{0.90}{0.00}
    & \numwild{0.51}{0.09} \\
    
    \midrule
    \rowcolor[HTML]{DFF1EC} 
    \multicolumn{12}{c}{Open-Source Models} \\
    \midrule
    
    Qwen2.5-VL-72B~\cite{bai2023qwen}
    & \numwild{0.18}{0.13} & \numwild{0.44}{0.13}
    & \numwild{0.72}{0.22} & \numwild{0.75}{0.27} & \numwild{0.73}{0.24}
    & \numwild{1.00}{0.00} & \numwild{0.50}{0.00} & \numwild{0.00}{0.00} & \numwild{1.00}{0.00} & \numwild{0.33}{0.00}
    & \numwild{0.55}{0.09} \\
    
    Qwen2.5-VL-7B~\cite{bai2023qwen}
    & \numwild{0.13}{0.13} & \numwild{0.14}{0.13}
    & \numwild{0.28}{0.22} & \numwild{0.39}{0.27} & \numwild{0.34}{0.24}
    & \numwild{0.00}{0.00} & \numwild{0.00}{0.00} & \numwild{0.00}{0.00} & \numwild{0.75}{0.00} & \numwild{0.03}{0.00}
    & \numwild{0.19}{0.09} \\
    
    Qwen2.5-VL-3B~\cite{bai2023qwen}
    & \numwild{0.14}{0.13} & \numwild{0.14}{0.13}
    & \numwild{0.22}{0.22} & \numwild{0.27}{0.27} & \numwild{0.24}{0.24}
    & \numwild{0.00}{0.00} & \numwild{0.00}{0.00} & \numwild{0.00}{0.00} & \numwild{0.00}{0.00} & \numwild{0.00}{0.00}
    & \numwild{0.09}{0.09} \\
    
    VLM-R1-3B~\cite{shen2025vlm}
    & \numwild{0.13}{0.13} & \numwild{0.13}{0.13}
    & \numwild{0.31}{0.22} & \numwild{0.34}{0.27} & \numwild{0.33}{0.24}
    & \numwild{0.00}{0.00} & \numwild{0.00}{0.00} & \numwild{0.00}{0.00} & \numwild{0.00}{0.00} & \numwild{0.00}{0.00}
    & \numwild{0.10}{0.09} \\
    
    VAGEN~\cite{wang2025vagen}
    & \numwild{0.61}{0.13} & \numwild{0.71}{0.13}
    & \numwild{0.78}{0.22} & \numwild{0.80}{0.27} & \numwild{0.79}{0.24}
    & \numwild{1.00}{0.00} & \numwild{0.88}{0.00} & \numwild{0.88}{0.00} & \numwild{0.88}{0.00} & \numwild{0.50}{0.00}
    & \numwild{0.78}{0.09} \\
    
    \midrule
    \rowcolor[HTML]{DFF1EC} 
    \multicolumn{12}{c}{Ours} \\
    \midrule
    
    \textbf{AtlasVA}
    & \numref{0.79}{0.13} & \numref{0.83}{0.13}
    & \numref{0.85}{0.22} & \numref{0.87}{0.27} & \numref{0.86}{0.24}
    & \numref{1.00}{0.00} & \numref{1.00}{0.00} & \numref{1.00}{0.00} & \numref{1.00}{0.00} & \numref{1.00}{0.00}
    & \numref{0.93}{0.09} \\
    
    \bottomrule
    \end{tabular}
    }
    \end{minipage}%
    \hfill
    \begin{minipage}[c]{0.06\textwidth}
    \colorbarvertical
    \end{minipage}
    \caption{Performance comparison across different tasks. Cell shading indicates relative performance within each column, with {\setlength{\fboxsep}{1pt}\protect\colorbox{monte_carlo}{deeper color}} denoting higher success rates.}
    \label{tab:vlm_results}
\end{table*}

In this section, we evaluate AtlasVA across diverse 2D and 3D spatial benchmarks and investigate the following research questions:

\researchquestion{
Does visual skill memory enable a compact 3B-parameter VLM agent to surpass significantly larger proprietary models on spatially intensive tasks? (Sec.~\ref{sec:main_result})
}{rq:rq1}

\researchquestion{
Does atlas-grounded dense reward shaping mitigate the sparse-reward bottleneck and accelerate policy convergence in long-horizon spatial tasks? (Sec.~\ref{sec:main_result}, Sec.~\ref{sec:ablation})
}{rq:rq2}

\researchquestion{
Can spatial heatmaps be effectively bootstrapped from raw trajectory statistics alone, without reliance on external teacher LLMs? (Sec.~\ref{sec:ablation})
}{rq:rq3}

\researchquestion{
How critical is each layer of the three-layer visual skill memory, and does native visual grounding provide clear advantages over text-only representations? (Sec.~\ref{sec:ablation})
}{rq:rq4}

\subsection{Experimental Setup}

\textbf{Base Model and Optimization.} We adopt Qwen2.5-VL-3B-Instruct as the base vision-language model. AtlasVA is optimized via Proximal Policy Optimization (PPO) with Generalized Advantage Estimation (GAE). During the rollout phase, the agent executes multi-turn interactions with the environment. We employ a training batch size of 128, setting the learning rate to $1\times 10^{-6}$ and $1\times 10^{-5}$ for the actor and critic networks, respectively. Full optimization hyperparameters are listed in Appendix~\ref{sec:supp_hyperparameters}.

\textbf{Visual Skill System Configuration.} The Three-Layer Visual Skill Memory provides continuous spatial grounding. To prevent information leakage, all memory components (heatmaps, exemplars, and text skills) are evolved exclusively using trajectories from the \textit{training} environments, keeping validation sets strictly separated for zero-shot evaluation. Detailed configurations, including EMA decay rates, exemplar pool capacities, and memory pruning mechanics, are provided in Appendix~\ref{sec:supp_hyperparameters}.

\textbf{Datasets.} We evaluate AtlasVA across four diverse agentic benchmarks: Classic Grid Puzzles (\textsc{Sokoban}~\cite{wang2025vagen} and \textsc{FrozenLake}~\cite{towers2024gymnasium}), Embodied 3D \textsc{Navigation}~\cite{yang2025embodiedbench,kolve2017ai2}, and Detailed Object Manipulation (\textsc{PrimitiveSkill}~\cite{tao2024maniskill3}). \textsc{PrimitiveSkill}, first introduced by VAGEN~\cite{wang2025vagen}, evaluates multi-turn robotic control. In addition to its standard tasks (Place, Stack, Drawer, and Align), we introduce a new \textit{Swap} task, which requires the agent to exchange the positions of two cubes.

\subsection{Main Results}
\label{sec:main_result}
We evaluate AtlasVA across four diverse environments, ranging from 2D discrete grid-worlds (\textsc{Sokoban}, \textsc{FrozenLake}) to continuous 3D embodied navigation and 3D robotic manipulation (PrimitiveSkill via ManiSkill). Quantitative comparisons are detailed in Table~\ref{tab:vlm_results}.

\textbf{Overall Performance.} AtlasVA achieves state-of-the-art performance across all evaluated benchmarks with an average success rate of 0.93 (Table~\ref{tab:vlm_results}). Despite using a compact 3B-parameter base model, AtlasVA significantly outperforms substantially larger proprietary models, including GPT-5 (0.69) and o3 (0.71) (\rqref{rq:rq1}). Furthermore, it establishes a considerable margin over the strongest open-source baseline, VAGEN (0.78). These results demonstrate that integrating visual skill memory with dense reward shaping effectively overcomes the performance ceiling of traditional text-centric VLM agents.

\textbf{Efficacy in Spatial Reasoning.} AtlasVA demonstrates strong spatial reasoning capabilities in environments that require intensive geometric planning, such as Sokoban and FrozenLake. Baseline open-source models exhibit limited spatial comprehension; for instance, zero-shot Qwen2.5-VL-3B obtains a success rate of merely 0.14 in Sokoban. By contrast, AtlasVA elevates this performance to 0.79, outperforming even the proprietary GPT-5 (0.70). For zero-shot transfer to unseen validation layouts with novel topologies, AtlasVA relies on the dynamic heuristic branch ($M_{heuristic}$) of the heatmaps, which computes topological features (e.g., BFS fields and corner traps) on-the-fly for the new layout, ensuring robust generalization despite shifted entities. This substantial gain empirically validates that progressively evolving spatial heatmaps successfully embed critical geometric priors, overcoming the inherent limitations of pure text-based representations in spatial tasks.

\textbf{Robustness in 3D Navigation and 3D Robotic Manipulation.} AtlasVA maintains robust execution across complex, continuous 3D tasks such as embodied Navigation and PrimitiveSkill robotic manipulation. On the 3D Navigation benchmark, AtlasVA yields an average success rate of 0.86 across all sub-tasks. In the 3D PrimitiveSkill manipulation domain (Place, Stack, Drawer, Align, Swap via ManiSkill), the method achieves a perfect 1.00 success rate across all five categories. This starkly contrasts with existing baselines, which frequently fail on complex manipulations; for instance, on the Swap task, Qwen2.5-VL-72B and GPT-5 achieve success rates of only 0.33 and 0.55, respectively. We attribute this robustness to the integration of contrasting visual exemplars, which effectively regularizes the policy and prevents the agent from revisiting historical failure modes during extended execution.

\textbf{Learning Efficiency and Convergence.} Beyond final task success, AtlasVA significantly accelerates training convergence compared to a text-centric baseline (Figure~\ref{fig:curve}). The baseline, relying solely on textual rules (Layer 3), struggles to exceed a 0.25 success rate in the 2D Sokoban environment due to severe modality mismatch and sparse feedback. In contrast, AtlasVA rapidly climbs to approximately 0.80 success within 140 training steps. Similarly, on the 3D PrimitiveSkill benchmark, AtlasVA quickly converges to a perfect success rate, whereas the baseline plateaus at roughly 0.60. This accelerated learning curve confirms that our teacher-free visual skill memory and dense visual reward shaping provide immediate, spatially grounded feedback, effectively solving the credit assignment problem and improving sample efficiency (\rqref{rq:rq2}).

\subsection{Ablation Studies}
\label{sec:ablation}

\begin{figure}[t]
    \centering
    \vspace{-8pt}

    \begin{minipage}[t]{0.48\linewidth}
        \centering
        \includegraphics[width=\linewidth]{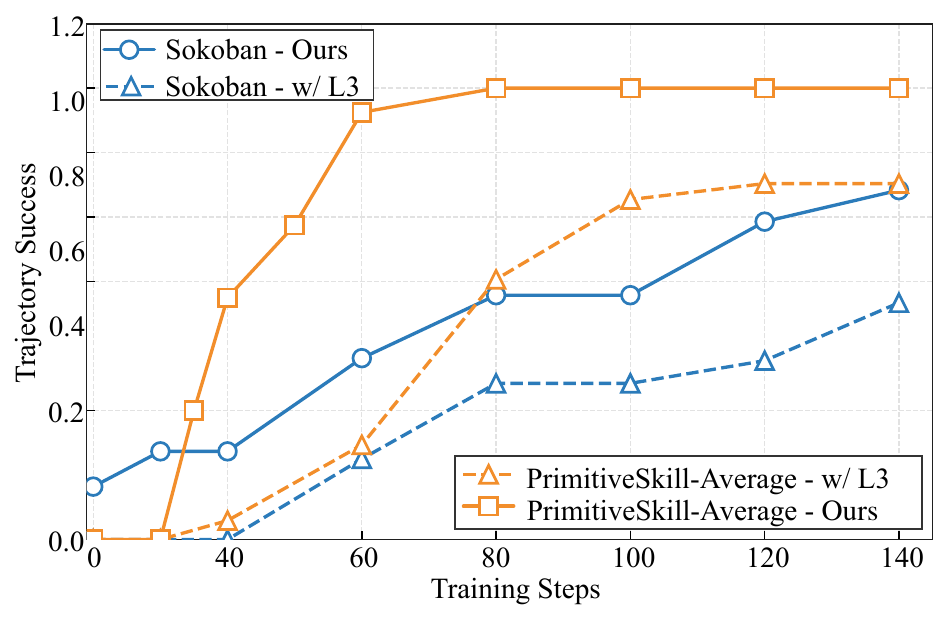}
        \vspace{-18pt}
        \captionof{figure}{\textbf{Learning efficiency.} AtlasVA (Ours) achieves faster convergence and higher task success than the text-only baseline (w/ L3), proving the efficiency of visual feedback.}
        \label{fig:curve}
    \end{minipage}
    \hfill
    \begin{minipage}[t]{0.48\linewidth}
        \centering
        \includegraphics[width=0.9\linewidth]{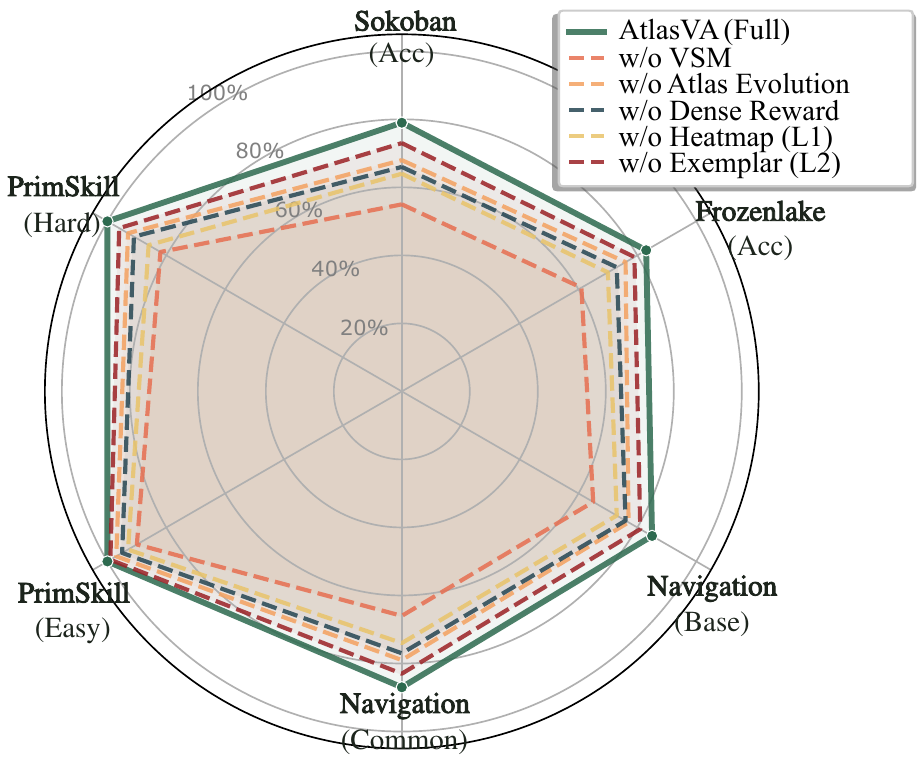}
        \vspace{-2pt}
        \captionof{figure}{\textbf{Component ablation.} Removing the visual skill memory (VSM), atlas evolution, or dense reward shaping consistently degrades performance across all spatial benchmarks.}
        \label{fig:radar_ablation}
    \end{minipage}

    \vspace{-6pt}
\end{figure}

\textbf{Diagnostic Analysis.} To evaluate AtlasVA's core components (Figure~\ref{fig:radar_ablation}), we analyze five ablations to address our research questions. \textit{Visual Skill Memory:} Removing the entire visual hierarchy (w/o VSM) severely degrades performance on spatial puzzles (Sokoban, FrozenLake), confirming that compressing geometric constraints into text causes fatal information loss and validating the necessity of native visual grounding (\rqref{rq:rq4}). Furthermore, omitting either topological maps (w/o Heatmap) or episodic retrieval (w/o Exemplar) hurts performance, proving both layers are required for optimal execution. \textit{Atlas Evolution:} Disabling off-policy updates (w/o Atlas Evolution) restricts the agent to static rules. The resulting major regression across all tasks proves that spatial heatmaps can be effectively bootstrapped from raw statistics alone, without external teacher LLMs (\rqref{rq:rq3}). \textit{Dense Reward Shaping:} A variant trained with only sparse feedback (w/o Dense Reward) struggles to escape local optima in long-horizon tasks. This confirms that translating the visual atlas into dense shaping signals is essential to mitigate the sparse-reward bottleneck and accelerate convergence (\rqref{rq:rq2}).

\begin{figure}[htbp]
  \centering
  \vspace{-5pt}
  \includegraphics[width=0.8\linewidth]{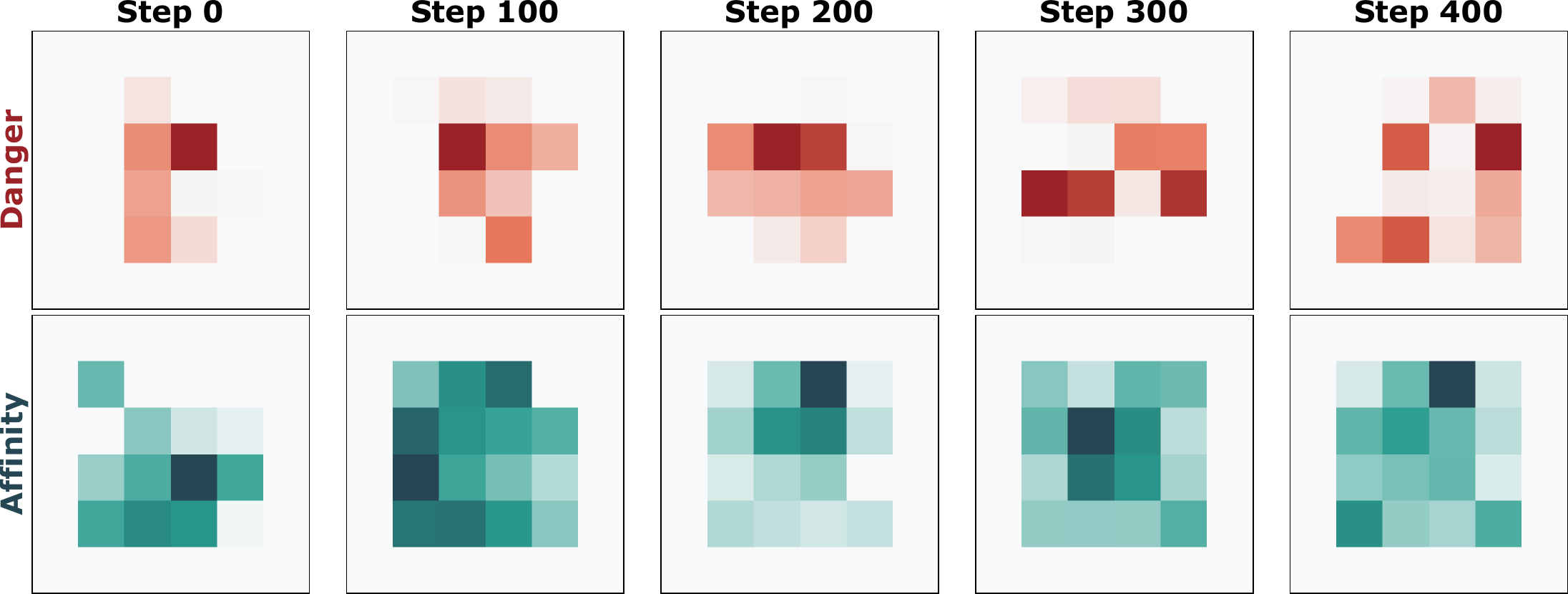}
  \caption{\textbf{Evolution of Spatial Heatmaps (Layer 1).} From Step 0 to Step 400, through pure environment interaction, the heatmaps learn to encode structural hazards (Danger, red) and sub-goal paths (Affinity, green).}
  \label{fig:heatmap_frames}
\end{figure}

\textbf{Evolution of Spatial Heatmaps.} Visualizing the progression of spatial heatmaps confirms that AtlasVA effectively extracts geometric priors through pure environment interaction (Figure~\ref{fig:heatmap_frames}). At Step 0, the initial heatmaps contain no spatial information. As training progresses via EMA updates, the heatmaps rapidly capture meaningful topological structures. By Step 200, the \textit{Danger} map (top row) explicitly highlights structural hazards and dead-ends, while the \textit{Affinity} map (bottom row) traces traversable sub-goal paths. This qualitative progression directly illustrates how the framework distills 2D layouts into visual priors without relying on external language model supervision.

\begin{figure}[htbp]
  \centering
  \setlength{\tabcolsep}{0pt}
  \vspace{-5pt}
  \begin{minipage}{0.48\linewidth}
    \centering
    \includegraphics[width=\linewidth]{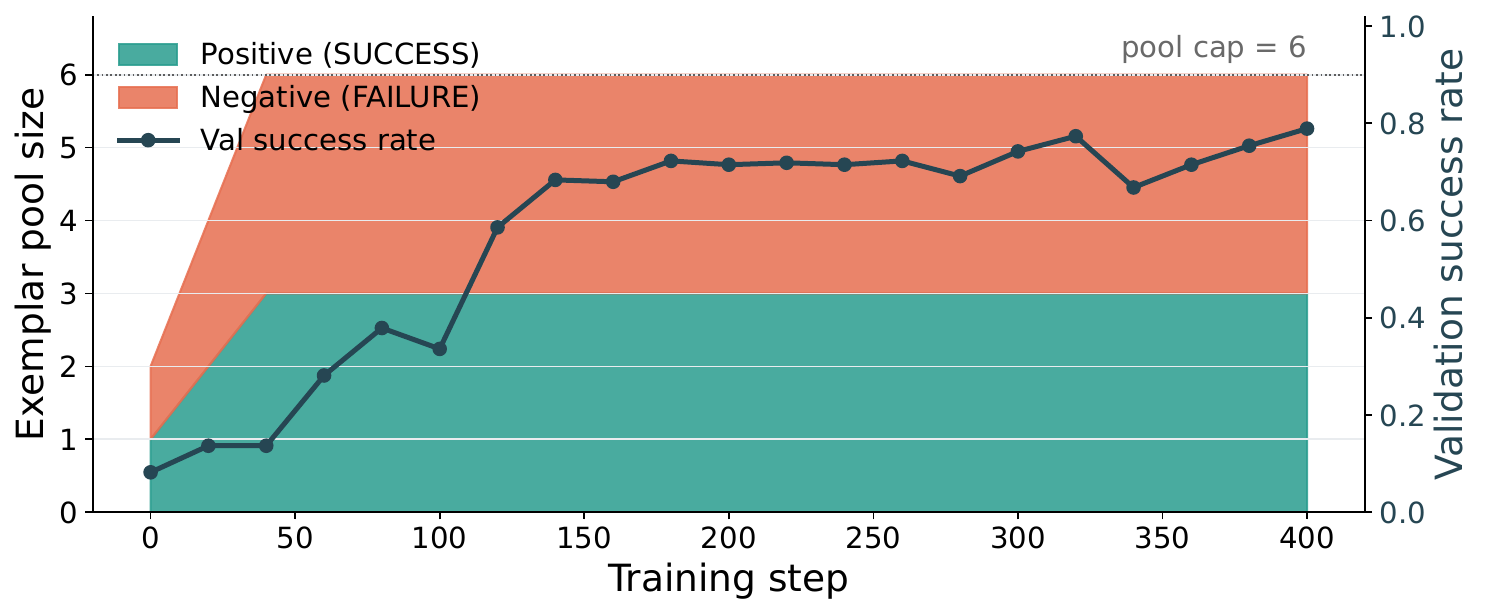}
    \caption{\textbf{Exemplar pool capacity.} The accumulation of visual exemplars coincides directly with a rapid increase in validation success rate.}
    \label{fig:exemplar_pool_A}
  \end{minipage}
  \hfill
  \begin{minipage}{0.48\linewidth}
    \centering
    \includegraphics[width=\linewidth]{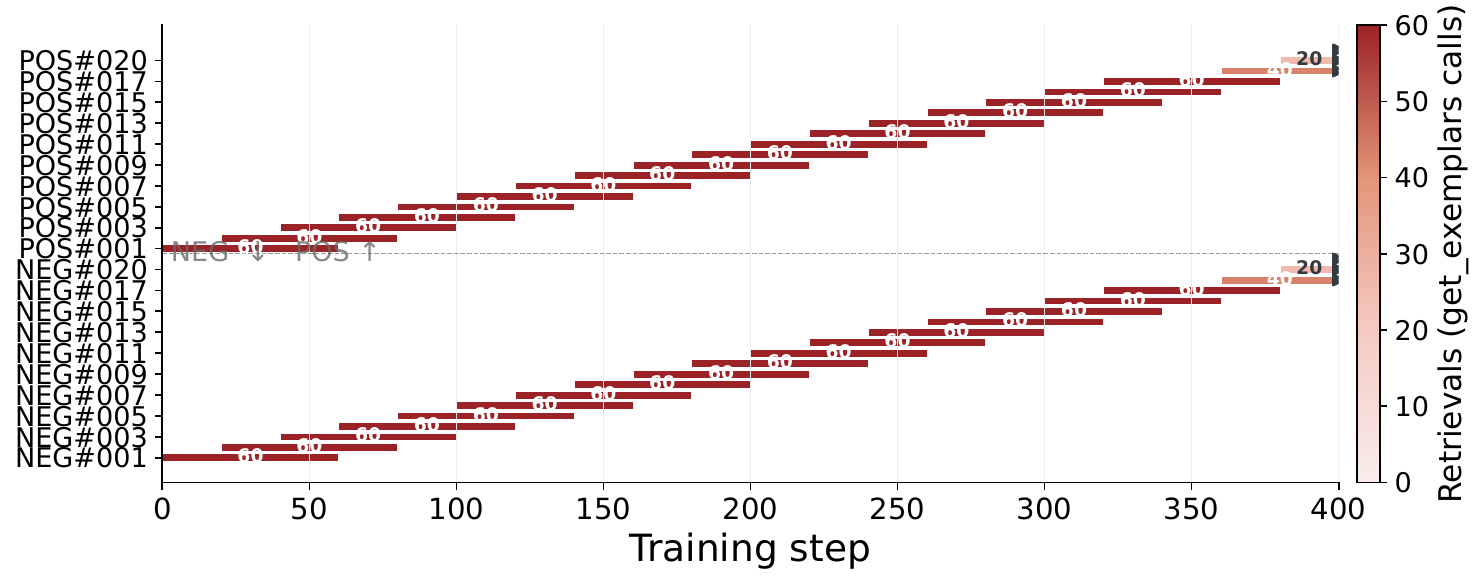}
    \caption{\textbf{Lifecycle of individual exemplars.} The pool continuously evicts older observations in favor of newer ones, providing dynamically updated visual context.}
    \label{fig:exemplar_pool_B}
  \end{minipage}
\end{figure}

\textbf{Dynamics of the Visual Exemplar Pool.} The visual exemplar pool dynamically updates to maintain highly relevant context throughout training, while keeping the prompt size strictly bounded (Figures~\ref{fig:exemplar_pool_A} and~\ref{fig:exemplar_pool_B}). Our implementation caps the pool at 6 exemplars, which it populates within the first 40 training steps (Figure~\ref{fig:exemplar_pool_A}). The rapid establishment of this pool directly correlates with an initial spike in the validation success rate from near zero to over 70\%, demonstrating the immediate benefit of visual references. Furthermore, Figure~\ref{fig:exemplar_pool_B} illustrates the continuous eviction mechanism: early exemplars (e.g., POS\#001) are gradually replaced by newer observations (e.g., POS\#020) as the agent explores novel states. The high retrieval frequency of active exemplars (indicated by shading intensity) confirms that the agent actively consults this rolling buffer, ensuring that the provided visual context matches its current exploration frontier.

\begin{figure}[htbp]
  \centering
  \includegraphics[width=1.0\linewidth]{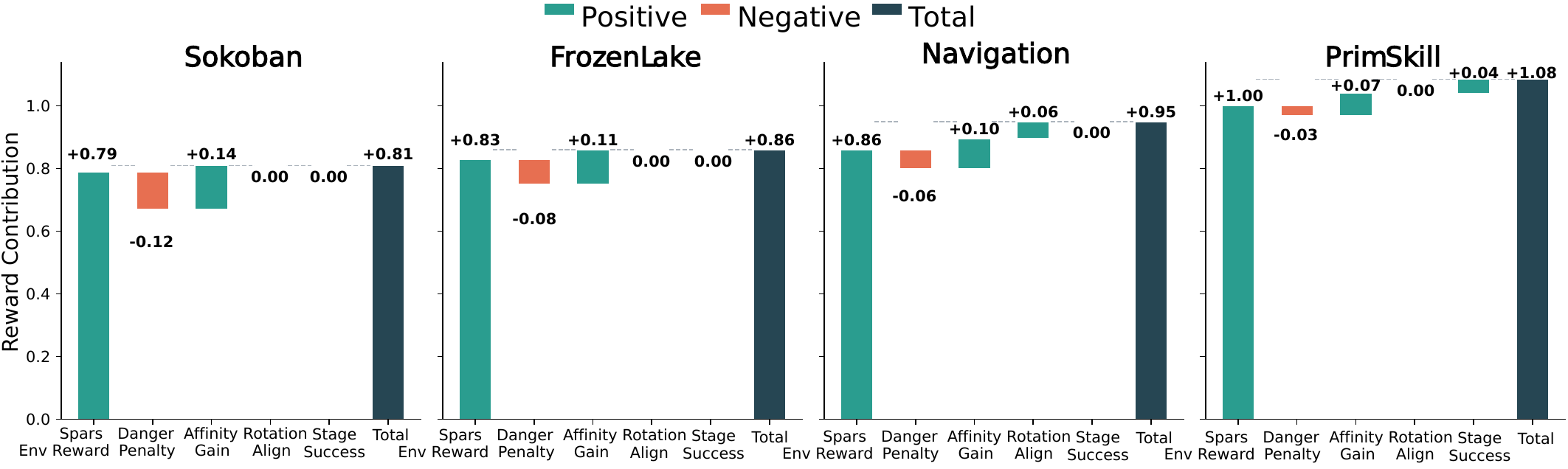}
  \caption{\textbf{Reward waterfall analysis.} The composition of the total reward demonstrates that the Danger Penalty (orange) and Affinity Gain (green) provide continuous step-level signals, effectively supplementing the sparse environment rewards.}
  \label{fig:reward_waterfall}
\end{figure}

\textbf{Impact of Dense Visual Reward Shaping.} Our Atlas-Grounded Dense Reward Shaping successfully converts sparse environment feedback ($r \in \{0, 1\}$), which provides inadequate gradients for long-horizon tasks, into a continuous optimization signal. As shown in Figure~\ref{fig:reward_waterfall}, our visual potential function explicitly supplements the sparse rewards: a \textbf{Danger Penalty} heavily discourages approaching learned hazards, while an \textbf{Affinity Gain} rewards progress toward sub-goals. This waterfall analysis verifies that translating 2D spatial heatmaps into a continuous potential field provides the dense guidance necessary to mitigate the credit assignment problem in sparse-reward settings.

\section{Conclusion and Limitations}
\label{sec:conclusion}

We presented \textbf{AtlasVA}, a native Visual Skill Memory that resolves spatial blindness and modality mismatch in VLM agents. By bootstrapping geometric priors directly from trajectories, AtlasVA enables continuous visual reward shaping without external teacher supervision. A compact 3B-parameter AtlasVA significantly outperforms larger models like GPT-5 on diverse spatial benchmarks. A primary limitation is the projection of table-top 3D manipulation into 2.5D visual priors; scaling this to highly occluded, ego-centric 3D robotics remains critical future work.


\newpage
\bibliographystyle{splncs04}
\bibliography{main}
\newpage
\beginappendix
\appendix

\section{Algorithmic and Mathematical Framework}
\label{sec:supp_math_algo}

While Section~\ref{sec:AtlasVA} introduced the conceptual framework of AtlasVA, including the Visual Skill Memory and dense reward shaping, this section provides the formal mathematical definitions and algorithmic flow of how these components fundamentally alter the underlying decision-making process. Specifically, we detail how the standard Partially Observable Markov Decision Process (POMDP) is augmented with an evolving observation space and a dynamic potential-based reward function.

\subsection{Standard POMDP vs. Augmented Observation Space}
A standard multi-turn VLM agentic task is typically defined as a POMDP tuple $\langle \mathcal{S}, \mathcal{A}, \mathcal{O}, \mathcal{T}, \mathcal{E}, \mathcal{R}, \gamma \rangle$, where:
\begin{itemize}
    \item $\mathcal{S}$ is the set of true unobservable environment states.
    \item $\mathcal{O}$ is the raw observation space (e.g., a first-person or top-down RGB image $o_t$).
    \item $\mathcal{R}(s_t, a_t) \to \{0, 1\}$ is the sparse environment reward.
\end{itemize}

In standard approaches (e.g., VAGEN), the VLM policy $\pi_\theta(a_t | o_{\le t})$ conditions only on the raw visual input $o_t$. To mitigate the spatial blindness discussed in Section~\ref{subsec:vsm}, AtlasVA explicitly expands the standard observation space. At any time step $t$ in episode $k$, the policy is conditioned on an augmented observation tuple $\tilde{o}_t$:
\begin{equation}
    \tilde{o}_t = \langle o_t, \mathcal{M}_{k}^{(heatmap)}, \mathcal{M}_{k}^{(exemplar)}, \mathcal{M}_{k}^{(text)} \rangle
\end{equation}
where $\mathcal{M}_{k}$ denotes the state of the three-layer Visual Skill Memory. This formalization ensures that non-local spatial priors (e.g., distant deadlock zones) become a native part of the state representation rather than relying on error-prone textual coordinate inference.

\subsection{Formalization of Potential-Based Dense Reward}
As introduced in Section~\ref{subsec:reward}, AtlasVA converts the highly sparse environment reward $r^{env}_t$ into a dense, per-step reward $\tilde{r}_t$. Here we provide the formal definition of the shaping term, inspired by potential-based reward shaping~\cite{ng1999policy} on the affinity branch and a heuristic safety constraint on the danger branch.

We define a visual potential field $\Phi(o_t)$ derived directly from the spatial heatmap $\mathcal{M}_{k}^{(heatmap)}$. The shaped reward is formulated as:
\begin{equation}
    \tilde{r}_t = r^{env}_t + F(o_t, a_t, o_{t+1})
\end{equation}
where the shaping term $F$ combines an affinity potential difference and a danger penalty:
\begin{equation}
    F(o_t, a_t, o_{t+1}) = \big[\Phi_{\text{affinity}}(o_{t+1}) - \Phi_{\text{affinity}}(o_t)\big] - \beta \cdot \mathbb{I}(\text{enters\_danger}(o_{t+1}))
\end{equation}
where:
\begin{itemize}
    \item $\Phi_{\text{affinity}}(o_t)$ evaluates the agent's current position on the affinity heatmap, whose base layer is a BFS distance field over reachable cells. The per-step difference $\Phi(o_{t+1}) - \Phi(o_t)$ therefore yields a non-trivial gradient at every state rather than only at terminals.
    \item $\mathbb{I}(\text{enters\_danger})$ triggers a danger penalty when the next observation falls into a region identified as a deadlock or fatal trap. We treat this term as a heuristic safety constraint that is intentionally non-potential-based: it deliberately reshapes the optimal policy to favor safer trajectories over hazardous shortcuts.
    \item $\beta$ is a scaling coefficient for the penalty.
\end{itemize}
The potential $\Phi_{\text{affinity}}$ blends static layout heuristics (BFS distance field) with EMA-accumulated trajectory statistics, providing a continuous visual gradient that transforms the sparse MDP into a dense optimization landscape.

\subsection{Off-Policy Memory Evolution Dynamics}
Traditional RL environments maintain static transition dynamics and reward functions. However, as conceptually described in Section~\ref{subsec:evolution}, AtlasVA operates as a dynamically evolving system. 

We mathematically define this evolution as an off-policy update loop. At the end of a rollout batch (comprising trajectories $\tau$), the spatial priors are updated via an Exponential Moving Average (EMA) mechanism:
\begin{equation}
    \mathcal{M}_{k+1}^{(heatmap)} \leftarrow \alpha \mathcal{M}_{k}^{(heatmap)} + (1-\alpha) \cdot \text{Extract}( \tau_k )
\end{equation}
where $\alpha$ is the EMA decay rate, and $\text{Extract}(\cdot)$ is a non-parametric mapping function. Specifically, $\text{Extract}$ aggregates terminal failure positions into the danger map and normalized state visitation frequencies of successful paths into the affinity map. These trajectory-driven statistics are then blended with the static layout heuristics $M_{heuristic}$ described in Section~\ref{subsec:evolution} (BFS distance field for affinity; corner / wall-adjacency / hole-neighborhood attenuation for danger), so that the final maps retain a dense spatial gradient at every cell even early in training. This formalizes a self-bootstrapping cycle: as the policy $\pi_\theta$ improves, the generated trajectories $\tau$ yield higher-quality updates to $\mathcal{M}$, which in turn provides more accurate potential functions $\Phi$ and augmented observations $\tilde{o}_t$ for subsequent policy optimization.

\subsection{Algorithmic Formalization of AtlasVA Evolution}
\label{sec:supp_algorithm}

To avoid redundancy with standard multi-turn PPO loops (which are well-documented in prior frameworks such as VAGEN), Algorithm~\ref{alg:atlasva} specifically formalizes the core mathematical contributions of this work: the \textbf{Teacher-Free Visual Atlas Evolution} and the \textbf{Atlas-Grounded Visual Reward Shaping}. It details how the visual priors are extracted from raw trajectory statistics and converted into dynamic potential functions.

\begin{algorithm}[!h]
\caption{Teacher-Free Visual Atlas Evolution and Dense Reward Shaping}
\label{alg:atlasva}
\begin{algorithmic}[1]
\STATE \textbf{Input:} Rollout buffer $\mathcal{B}$ from the current PPO epoch, containing spatial coordinates $\mathbf{p}_t$, observations $o_t$, actions $a_t$, and sparse rewards $r^{env}_t$.
\STATE \textbf{State Variables:} Historical Danger Map $M_{danger}$, Affinity Map $M_{affinity}$, Exemplar Pool $\mathcal{E}_{vis}$.
\STATE \textbf{Hyperparameters:} EMA decay $\alpha$, reward scaling $\beta, \gamma$.
\STATE
\STATE \textcolor{gray}{\# Phase 1: Atlas-Grounded Visual Reward Shaping (Online)}
\FOR{each transition $(\mathbf{p}_t, a_t, \mathbf{p}_{t+1})$ in $\mathcal{B}$}
    \STATE $\Phi_{\text{affinity}} \leftarrow M_{affinity}(\mathbf{p}_{t+1})$
    \STATE $r_{affinity} \leftarrow \Phi_{\text{affinity}} - M_{affinity}(\mathbf{p}_t)$
    \STATE $r_{danger} \leftarrow -\beta \cdot M_{danger}(\mathbf{p}_{t+1})$
    \STATE $\tilde{r}_t \leftarrow r^{env}_t + r_{affinity} + r_{danger}$ \quad \textcolor{gray}{\# Construct dense visual reward}
    \STATE Replace $r^{env}_t$ with visual shaped reward $\tilde{r}_t$ in $\mathcal{B}$
\ENDFOR
\STATE \textbf{Execute Standard PPO Update} for policy $\pi_\theta$ using the dense visual rewards $\tilde{r}_t$ in $\mathcal{B}$
\STATE
\STATE \textcolor{gray}{\# Phase 2: Teacher-Free Atlas Evolution (Offline)}
\STATE Extract failed trajectories $\mathcal{T}_{fail}$ and successful trajectories $\mathcal{T}_{succ}$ from $\mathcal{B}$
\STATE \textcolor{gray}{\# 2.1 Danger Map Evolution}
\STATE Initialize batch danger map $M_{batch}^{danger} \leftarrow \mathbf{0}$
\FOR{each trajectory $\tau \in \mathcal{T}_{fail}$}
    \STATE $\mathbf{p}_{terminal} \leftarrow$ terminal coordinate of $\tau$
    \STATE $M_{batch}^{danger}(\mathbf{p}_{terminal}) \mathrel{+}= 1 / |\mathcal{T}_{fail}|$
\ENDFOR
\STATE $M_{danger} \leftarrow \alpha M_{danger} + (1-\alpha) M_{batch}^{danger}$
\STATE \textcolor{gray}{\# 2.2 Affinity Map Evolution}
\STATE Initialize batch affinity map $M_{batch}^{affinity} \leftarrow \mathbf{0}$
\FOR{each trajectory $\tau \in \mathcal{T}_{succ}$}
    \FOR{each visited coordinate $\mathbf{p} \in \tau$}
        \STATE $M_{batch}^{affinity}(\mathbf{p}) \mathrel{+}= 1 / (\text{length of } \tau \times |\mathcal{T}_{succ}|)$
    \ENDFOR
\ENDFOR
\STATE $M_{affinity} \leftarrow \alpha M_{affinity} + (1-\alpha) M_{batch}^{affinity}$
\STATE \textcolor{gray}{\# 2.3 Visual Exemplar Pool Update}
\STATE Identify critical inflection keyframes from $\mathcal{T}_{fail}$ and $\mathcal{T}_{succ}$
\STATE Update $\mathcal{E}_{vis}$ via DINOv2 cosine distance matching and FIFO eviction
\STATE
\STATE \textbf{Return:} Updated Visual Skill Memory $\{M_{danger}, M_{affinity}, \mathcal{E}_{vis}\}$
\end{algorithmic}
\end{algorithm}

\section{Prompt Template}
\label{sec:supp_prompt_template}

\definecolor{myboxbg}{HTML}{EFF5F6}
\definecolor{myboxframe}{HTML}{82B1B3} 
\definecolor{myred}{HTML}{C53A32}

\begin{figure}[h]
\centering
\begin{tcolorbox}[
colback=myboxbg,
colframe=myboxframe,
boxrule=0.8pt,
arc=1pt,
left=4pt,
right=4pt,
top=3pt,
bottom=3pt
]

\small\ttfamily
[System] Sokoban rules and output format ...\\
\textcolor{teal}{\#\# Spatial Skill Maps}\\
\quad Danger zones (red): \textlangle image\textrangle\\
\quad Goal affinity (green): \textlangle image\textrangle\\
\textcolor{teal}{\#\# Visual Exemplars} \textit{(optional, when pool is non-empty)}\\
\quad Positive cases: \textlangle image\textrangle, \textlangle image\textrangle, ...\\
\quad Negative cases: \textlangle image\textrangle, \textlangle image\textrangle, ...\\
\textcolor{teal}{\#\# Learned Principles}\\
\quad \#\#\# General Principles / Push Strategies / Mistakes to Avoid ...\\
{} \\
{}[User] [Initial Observation]: \textlangle image\textrangle\\
\quad Decide your next action(s).
\end{tcolorbox}
\caption{Prompt template of AtlasVA. The three VSM layers (L1/L2/L3) are injected at fixed anchor positions, followed by the current observation in the user turn.}
\label{lst:prompt_template}
\end{figure}

\section{Implementation and Environmental Details}
\label{sec:supp_implementation}

This section provides the construction details of the Visual Skill Memory (VSM) and the exact reward assignment hyperparameters used across our evaluated environments.

\subsection{Layer 1: Spatial Heatmaps (Dense Perceptual Priors)}
While Section~\ref{subsec:vsm} introduces the concept of rendering $M_{danger}$ and $M_{affinity}$ as RGB heatmaps, here we detail the precise tensor-to-image mapping pipeline used to bridge the discrete grid statistics and the VLM's continuous visual encoder.

\textbf{Construction \& Rendering.} 
The environment layout is mapped to a discrete 2D grid $\mathbf{p} = (x, y)$. During the EMA evolution phase, the raw aggregated statistics $M^{danger}(\mathbf{p})$ and $M^{affinity}(\mathbf{p})$ are stored as floating-point tensors in $[0, 1]$. To render these into visual tokens without losing precision or confusing the VLM's pre-trained color semantics:
\begin{itemize}
    \item \textbf{Alpha-Channel Gradient Mapping:} We apply a strict linear colormap. The danger tensor is mapped to a pure red channel (\texttt{R=255, G=0, B=0}) with the alpha (opacity) channel directly proportional to the danger value. The affinity tensor is identically mapped to a pure green channel (\texttt{R=0, G=255, B=0}). 
    \item \textbf{Tokenization Strategy:} As shown in the prompt template (Appendix~\ref{sec:supp_prompt_template}), we inject these rendered heatmaps as standalone auxiliary image tokens (i.e., \texttt{<image>}) rather than directly alpha-blending them over the current observation $o_t$. Empirical testing revealed that direct blending often corrupts the VLM's recognition of small foreground interactive objects. By providing them as separate spatial reference maps, the VLM natively learns to cross-attend between the clean observation and the semantic heatmaps.
\end{itemize}

\subsection{Layer 2: Visual Exemplars (Episodic Context)}
To compensate for the loss of temporal causality in the aggregated heatmaps, Layer 2 maintains an episodic memory of critical keyframes ($\mathcal{E}_{vis}$). Here we detail the dynamic retrieval mechanism.

\textbf{Construction and Retrieval.} 
The exemplar pool is dynamically populated during the off-policy evolution phase. We filter historical trajectories to extract ``inflection point'' frames—e.g., the exact frame immediately preceding an irreversible deadlock or a critical sub-goal completion.

During inference, we deploy a dense retrieval mechanism. Given the current observation $o_t$, we retrieve the top-$k$ most structurally relevant frames from the exemplar pool:
\begin{equation}
    \mathcal{E}_t = \underset{e_i \in \mathcal{E}}{\mathrm{arg\,max}} \, \text{CosSim}(f_{\text{DINO}}(o_t), f_{\text{DINO}}(e_i))
\end{equation}
where $f_{\text{DINO}}(\cdot)$ denotes the feature embeddings extracted by a frozen DINOv2 encoder. DINOv2 is chosen over CLIP because its self-supervised objective yields superior patch-level spatial correspondences, which is crucial for 2D topological matching in environments like Sokoban. The retrieved frames are then partitioned into ``Positive Cases'' and ``Negative Cases'' and appended to the prompt, enabling zero-shot visual pattern matching.

\subsection{Layer 3: Symbolic Text Skills (Semantic Grounding)}
\label{sec:supp_layer3}
While the final prompt structure (General Principles, Push Strategies, Mistakes to Avoid) is illustrated in Listing~\ref{lst:prompt_template}, we clarify that these symbolic rules do not require any proprietary Teacher LLM (e.g., GPT-4) to generate or summarize. Instead, $\mathcal{S}_{text}$ is constructed directly from the environment's underlying rulebook and task descriptions. 

This static text acts as the cognitive scaffolding that binds the VSM together: the textual rules dictate the foundational logic (e.g., what constitutes a valid push), while the self-evolved heatmaps (Layer 1) and retrieved exemplars (Layer 2) visually ground \textit{where} and \textit{how} that logic applies in the current continuous pixel space. By sourcing text skills directly from environment specifications, AtlasVA remains completely teacher-free.

\subsection{Environment Details and Reward Assignment}
\label{sec:supp_env_rewards}

To facilitate reproducibility, we detail the exact reward assignment hyper-parameters across all evaluated environments in Table~\ref{tab:reward_params}. 

While the sparse environment rewards (Success and Failure) provide the terminal objective, AtlasVA introduces two new dense visual reward scaling coefficients: $\lambda_{danger}$ (controlling the penalty for entering deadlock regions identified by the Danger Map) and $\lambda_{affinity}$ (controlling the step-wise gain for moving towards goal regions identified by the Affinity Map). We empirically found that tasks requiring high-precision, long-horizon coordinate manipulation (e.g., PrimitiveSkill) benefit from stronger visual guidance ($\lambda_{danger} = 0.3, \lambda_{affinity} = 0.3$) compared to pure grid-world planning tasks like Sokoban ($\lambda_{danger} = 0.05, \lambda_{affinity} = 0.05$).

\textbf{Clarification on Reward Sparsity and Auxiliary Components.} As formulated in Section~\ref{subsec:problem}, the underlying environment dynamics strictly provide sparse binary rewards $\mathcal{R} \in \{0, 1\}$. However, to successfully train VLM agents, our framework incorporates three auxiliary components (as observed in the reward waterfall analysis):
\begin{itemize}
    \item \textbf{Format Penalty:} A system-level constraint applied when the VLM generates malformed actions or invalid JSON. This is a syntactic regularizer independent of the spatial environment dynamics.
    \item \textbf{Rotation Align \& Stage Success:} To accommodate complex 3D environments, the visual shaping reward $r_{visual}$ is extended with task-specific spatial heuristics. Specifically, a rotation alignment bonus is added for 3D Navigation to encourage facing the goal, and a stage-success delta is folded into the Affinity Gain for multi-step PrimitiveSkill manipulations. These are extensions of our proposed dense shaping mechanism, preserving the strictly sparse nature of the base environment.
\end{itemize}

\begin{table}[htbp]
\centering
\caption{\textbf{Reward assignment parameters across evaluated environments.} Standard sparse rewards (Success/Failure) are heavily augmented by AtlasVA's continuous visual gradients, governed by $\lambda_{danger}$ and $\lambda_{affinity}$.}
\label{tab:reward_params}
\small
\setlength{\tabcolsep}{6pt}
\renewcommand{\arraystretch}{1.2}
\begin{tabular}{@{}lccccc@{}}
\toprule
\textbf{Environment} & \textbf{Success} & \textbf{Failure/Timeout} & \textbf{Format Penalty} & $\bm{\lambda_{danger}}$ \textbf{(Penalty)} & $\bm{\lambda_{affinity}}$ \textbf{(Gain)} \\ \midrule
Sokoban             & +1.0             & -0.1                     & -0.5                   & 0.05                             & 0.05                            \\
FrozenLake          & +1.0             & -0.1                     & -0.5                   & 0.05                             & 0.05                            \\
Navigation          & +1.0             & -0.1                     & -0.5                   & 0.1                              & 0.1                             \\
PrimitiveSkill      & +1.0             & 0.0                      & -0.5                   & 0.3                              & 0.3                             \\ \bottomrule
\end{tabular}
\end{table}

\subsection{GridState Abstraction and Simulator APIs}
\label{sec:supp_gridstate_api}

As discussed in Section~\ref{subsec:problem}, the \textit{GridState} abstraction $g_t$ extracts exact 2D/3D coordinates ($\mathbf{p}_t$) directly from the underlying simulator to construct the spatial heatmaps and compute dense visual rewards. Table~\ref{tab:gridstate_api} explicitly details the low-level simulator APIs accessed for each environment.

Crucially, this privileged state access is \textbf{strictly confined to the offline atlas evolution and reward computation pipeline during training}. At evaluation time, the VLM policy does not receive any of these low-level states. The policy's input consists entirely of the raw RGB observation $o_t$ and the rendered visual memory prompts (heatmaps and exemplars), ensuring that the zero-shot comparisons in our main results remain strictly fair and vision-based.

\begin{table}[htbp]
\centering
\caption{\textbf{Simulator APIs accessed by the GridState abstraction.} These privileged states are used exclusively for training-time memory evolution and reward shaping, not as policy inputs.}
\label{tab:gridstate_api}
\small
\setlength{\tabcolsep}{8pt}
\renewcommand{\arraystretch}{1.2}
\begin{tabular}{@{}ll@{}}
\toprule
\textbf{Environment} & \textbf{Accessed Simulator API / Internal State} \\ \midrule
Sokoban & \texttt{room\_state}, \texttt{room\_fixed}, \texttt{player\_position}, \texttt{boxes\_on\_target} \\
FrozenLake & \texttt{gym\_env.s} (agent index), \texttt{gym\_env.desc} (map layout) \\
Navigation & AI2-THOR metadata (agent poses, reachable positions, goal coordinates) \\
PrimitiveSkill & ManiSkill internal states (\texttt{\_handle\_info}, \texttt{last\_info}, object poses) \\ \bottomrule
\end{tabular}
\end{table}

\subsection{Optimization and Hyperparameter Configurations}
\label{sec:supp_hyperparameters}

In Table~\ref{tab:ppo_vsm_params}, we detail the unified hyperparameters used for the Proximal Policy Optimization (PPO) training and the Visual Skill Memory (VSM) configurations. All experiments were conducted using 8 $\times$ NVIDIA RTX 6000 Ada Generation GPUs. The VLM base model is Qwen2.5-VL-3B-Instruct. We utilize the vLLM engine for asynchronous rollout generation, enforcing strict maximum lengths for prompt and response tokens to accommodate the injected multimodal memory.

The VSM hyperparameters, including the EMA decay rate ($\lambda=0.85$) and the capacity of the visual exemplar pool (3 positive and 3 negative cases), were kept consistent across all environments to demonstrate the robustness and generalizability of the teacher-free evolution mechanism.

\begin{table}[htbp]
\centering
\caption{\textbf{Comprehensive hyperparameter configurations for PPO optimization and VSM components.}}
\label{tab:ppo_vsm_params}
\small
\setlength{\tabcolsep}{8pt}
\renewcommand{\arraystretch}{1.1}
\begin{tabular}{@{}llc@{}}
\toprule
\textbf{Category} & \textbf{Hyperparameter} & \textbf{Value} \\ \midrule
\multirow{6}{*}{\textbf{PPO \& GAE}} 
& Actor Learning Rate & $1 \times 10^{-6}$ \\
& Critic Learning Rate & $1 \times 10^{-5}$ \\
& Train Batch Size (Rollout) & 128 \\
& PPO Mini-Batch Size & 32 \\
& KL Control Coefficient & 0.0 \\
& Number of GPUs & 8 \\ \midrule
\multirow{4}{*}{\textbf{Rollout Engine}} 
& Inference Engine & vLLM (async) \\
& Max Batched Tokens & 16384 \\
& Max Prompt Length & 5000 - 8000 \\
& Max Response Length & 4000 \\ \midrule
\multirow{5}{*}{\textbf{Layer 1: Heatmap}} 
& EMA Decay Rate ($\alpha$) & 0.85 \\
& Render Cell Size (pixels) & 40 \\
& Active Field Types & Danger, Affinity \\ \midrule
\multirow{3}{*}{\textbf{Layer 2: Exemplar}} 
& Max Positive / Negative Pool & 3 / 3 \\
& Max Exemplars Injected & 4 \\
& Retrieval Metric & DINOv2 Cosine Sim. \\ \midrule
\multirow{3}{*}{\textbf{Layer 3: Text Skill}} 
& Top-$K$ Skills Injected & 3 (Grid), 4 (Swap) \\
& Pruning Frequency & Every 10 updates \\
\bottomrule
\end{tabular}
\end{table}

\begin{figure}[ht!]
    \centering
    \includegraphics[width=1.0\linewidth]{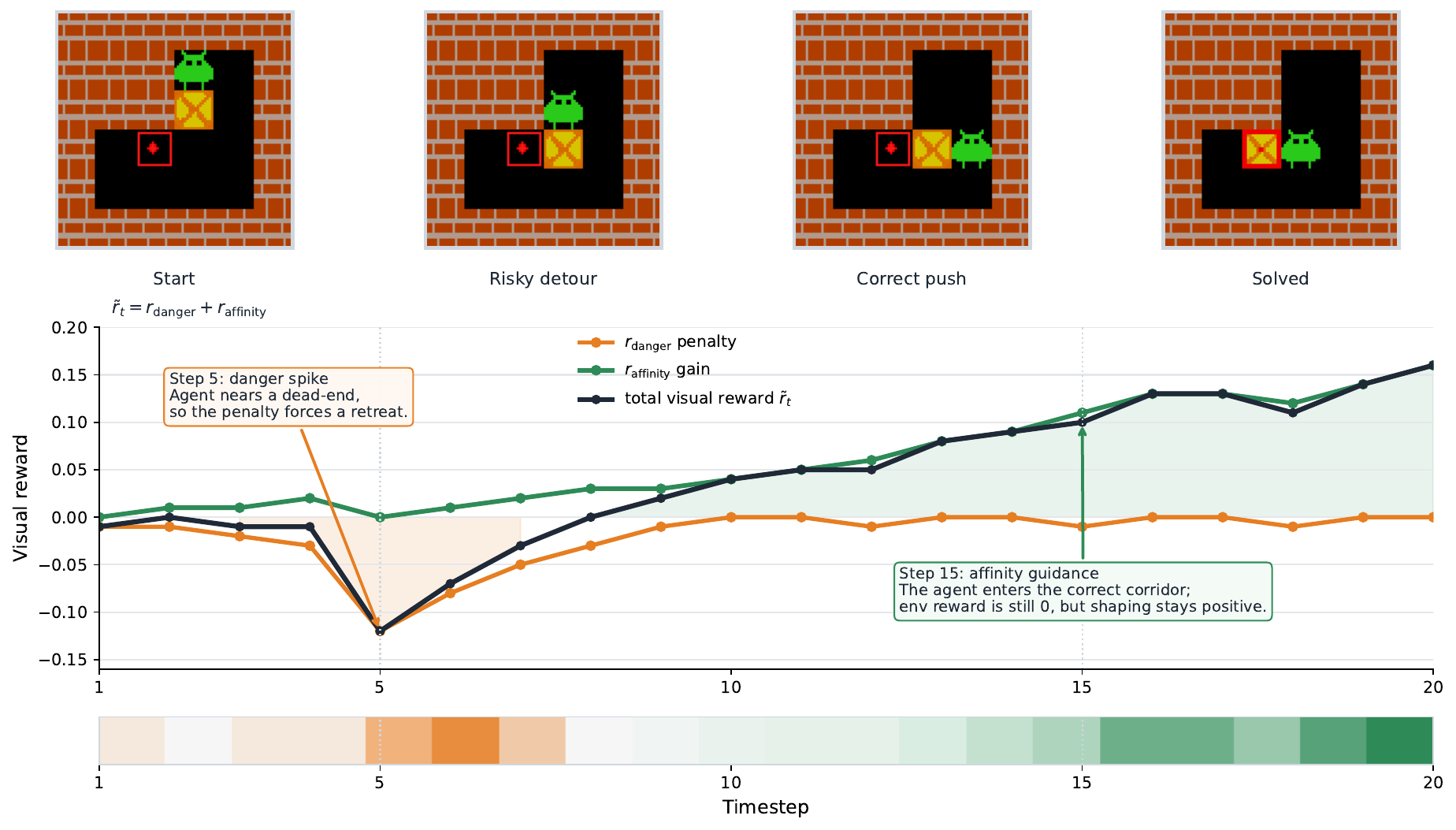}
    \caption{\textbf{Step-by-step visual reward dynamics in a Sokoban trajectory.} The plot expands on the reward composition discussed in Section~\ref{subsec:reward} and Figure~\ref{fig:reward_waterfall} by demonstrating how AtlasVA converts a single terminal sparse reward into a continuous optimization landscape over time. The Danger Penalty (orange) issues immediate corrective feedback when the agent approaches deadlocks, while the Affinity Gain (green) provides sustained positive signals as the agent navigates along viable sub-goal paths.}
    \label{fig:trajectory_reward_sokoban}
\end{figure}

\section{Additional Qualitative Analysis}
\label{sec:supp_qualitative_analysis}

\subsection{Projection of Continuous 3D Spaces into 2.5D Visual Priors}
\label{sec:supp_3d_projection}

A natural question arises when applying AtlasVA to continuous 3D embodied navigation and 3D robotic manipulation tasks (e.g., ManiSkill): how does a 2D visual atlas accommodate 3D physical spaces?

AtlasVA handles this through an elegant dimensionality-reduction abstraction layer (via the \texttt{GridState} module). For 3D Navigation tasks, the environment's continuous reachable surfaces are projected onto the X-Z plane, dynamically discretizing the 3D room into a 2D floor plan (with a resolution of 0.25 meters per cell). For 3D Robotic Manipulation (PrimitiveSkill), the continuous operating table is mapped to a localized 2.5D workspace grid, while the critical Z-axis (height) information is implicitly preserved as task metadata. 

During reward shaping, the agent's real-time continuous 3D coordinates are inversely mapped to this 2.5D heatmap. This allows AtlasVA to look up the evolving spatial potential and generate dense, coordinate-aligned gradients. Consequently, the agent executes complex 3D manipulations guided by highly interpretable 2.5D visual priors, significantly reducing the sample complexity typically associated with 3D continuous RL.

\begin{figure}[!ht]
  \centering
  \includegraphics[width=1.0\linewidth]{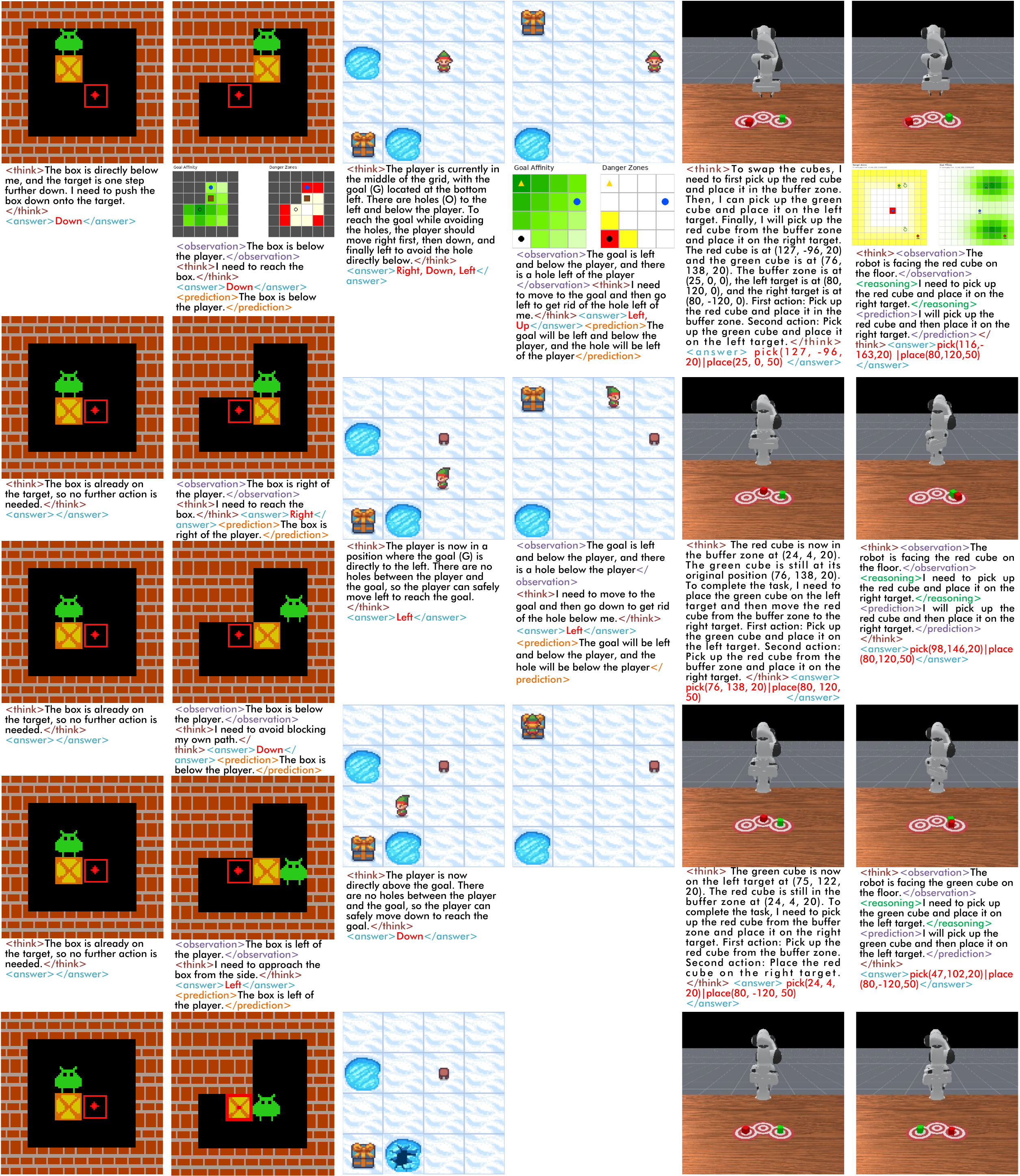}
  \caption{\textbf{Qualitative rollout comparison between Qwen2.5-VL-72B and AtlasVA.} Across spatially demanding tasks (Sokoban, FrozenLake, and PrimitiveSkill Swap), the 72B baseline struggles with geometric reasoning, often falling into deadlocks or failing long-horizon manipulations due to spatial blindness. In contrast, AtlasVA (3B) leverages injected spatial heatmaps (danger in red, affinity in green) to intuitively perceive topological hazards and sub-goals, successfully completing the tasks through precise, native visual grounding.}
  \label{fig:Comparison}
\end{figure}

\subsection{Visualization of Atlas-Grounded Visual Reward Shaping}
\label{sec:supp_reward_vis}

To provide a concrete understanding of how the visual potential fields alleviate the credit assignment problem in sparse-reward environments, we visualize the step-by-step dense reward dynamics of a successful Sokoban rollout in Figure~\ref{fig:trajectory_reward_sokoban}.

In this environment, the agent only receives a sparse extrinsic reward of $+1$ upon completing the entire task. Without intermediate feedback, VLM agents typically resort to random walks or fall into deadlocks. However, as illustrated by the trajectory curves, the proposed Atlas-Grounded Visual Reward Shaping mechanism provides a continuous, coordinate-specific gradient throughout the episode:

\begin{itemize}
    \item \textbf{Danger Penalties (Orange Curve):} When the agent takes an action that moves it toward an irreversible corner or wall trap, the visual potential field $\Phi_{danger}$ immediately induces a negative penalty. This provides a prompt corrective signal, forcing the agent to backtrack before a terminal failure occurs.
    \item \textbf{Affinity Gains (Green Curve):} Conversely, as the agent pushes the box along the structurally valid paths toward the goal manifold, the $\Phi_{affinity}$ potential yields sustained positive rewards. These continuous step-level gains guide the agent through the long-horizon task.
\end{itemize}

By seamlessly replacing sparse feedback with dense, spatially grounded visual gradients, AtlasVA effectively bridges the modality gap and stabilizes multi-turn RL optimization.

\subsection{Qualitative Rollout Comparison}
\label{sec:supp_qualitative}

To provide deeper intuition into how the Visual Skill Memory fundamentally alters the agent's spatial reasoning capabilities, we visualize complete rollout trajectories in Figure~\ref{fig:Comparison}. We compare our proposed AtlasVA against a strong baseline, Qwen2.5-VL-72B, across three spatially demanding environments: Sokoban, FrozenLake, and the PrimitiveSkill Swap task.

As illustrated in the baseline trajectories of Figure~\ref{fig:Comparison}, standard VLMs frequently exhibit ``spatial blindness.'' Despite having a massive parameter count and robust text reasoning capabilities, the 72B baseline struggles to ground abstract textual rules into the unannotated 2D geometric layout. In Sokoban and FrozenLake, this modality mismatch causes the baseline to push boxes into irreversible corners or step into topological traps. In the PrimitiveSkill Swap task, it fails to execute the precise, long-horizon coordinate manipulations required to complete the rearrangement.

Conversely, the AtlasVA trajectories demonstrate the power of multimodal spatial grounding. By natively injecting the self-evolved spatial heatmaps—where red regions highlight topological deadlocks (Danger) and green regions indicate optimal sub-goal paths (Affinity)—AtlasVA transforms complex geometric planning into intuitive visual pattern matching. Even when initialized from a compact 3B parameter model, AtlasVA successfully navigates these bottlenecks without relying on external text summaries, showcasing the immense sample efficiency and spatial awareness unlocked by our teacher-free visual memory.

\end{document}